\documentclass[10pt,twocolumn,letterpaper]{article}

\usepackage{iccv}
\usepackage{times}
\usepackage{epsfig}
\usepackage{graphicx}
\usepackage{amsmath}
\usepackage{amssymb}
\usepackage{booktabs}
\usepackage{mathrsfs} 

\usepackage{algorithm}
\usepackage{algorithmic}
\usepackage{verbatim}
\usepackage{color}
\usepackage{bbding}
\usepackage{bbm}
\usepackage{subfigure}
\usepackage{multirow}
\usepackage[accsupp]{axessibility}

\usepackage[pagebackref=true,breaklinks=true,letterpaper=true,colorlinks,bookmarks=false]{hyperref}

\newif\ifshowcomments
\showcommentstrue

\ifshowcomments
\newcommand{\TODO}[1]{{\color{red}{[TODO: #1]}}}

\newcommand{\revised}[1]{{\color[rgb]{0.2,0.7,0.2}{#1}}}

\newcommand{\xwhu}[1]{{\color[rgb]{0.5,0.2,0.7}{#1}}}

\else
\newcommand{\TODO}[1]{}
\newcommand{\revised}[1]{}
\fi

\usepackage[capitalize]{cleveref}
\crefname{section}{Sec.}{Secs.}
\Crefname{section}{Section}{Sections}
\Crefname{table}{Table}{Tables}
\crefname{table}{Tab.}{Tabs.}

\def\iccvPaperID{4899} 

\ificcvfinal\fi
\iccvfinalcopy

\begin{document}
\title{SILT: Shadow-aware Iterative Label Tuning for\\ Learning to Detect Shadows from Noisy Labels}
\author{Han Yang$^{1, \ast}$, Tianyu Wang$^{1,2}$\thanks{Joint first authors}\ , Xiaowei Hu$^{3,1, }$\thanks{Corresponding author (xwhu@cse.cuhk.edu.hk)}, and Chi-Wing Fu$^{1,2}$\vspace{1mm}\\
$^1$ The Chinese University of Hong Kong
$^2$ The Shun Hing Institute of Advanced Engineering\\
$^3$ Shanghai Artificial Intelligence Laboratory\\
}
\maketitle


\begin{abstract}
Existing shadow detection datasets often contain missing or mislabeled shadows, which can hinder the performance of deep learning models trained directly on such data. 
To address this issue, we propose SILT, the Shadow-aware Iterative Label Tuning framework, which explicitly considers noise in shadow labels and trains the deep model in a self-training manner.
Specifically, we incorporate strong data augmentations with shadow counterfeiting to help the network better recognize non-shadow regions and alleviate overfitting. 
We also devise a simple yet effective label tuning strategy with global-local fusion and shadow-aware filtering to encourage the network to make significant refinements on the noisy labels.
We evaluate the performance of SILT by relabeling the test set of the SBU~\cite{vicente2016large} dataset and conducting various experiments. 
Our results show that even a simple U-Net~\cite{ronneberger2015u} trained with SILT can outperform all state-of-the-art methods by a large margin. When trained on SBU / UCF~\cite{zhu2010learning} / ISTD~\cite{wang2018stacked}, our network can successfully reduce the Balanced Error Rate by 25.2\% / 36.9\% / 21.3\% over the best state-of-the-art method.

%
\end{abstract}


\section{Introduction}
\label{sec:intro}



\begin{figure}[t]
	\begin{minipage}[t]{\linewidth}
	\centering
		\includegraphics[width=0.92\linewidth]{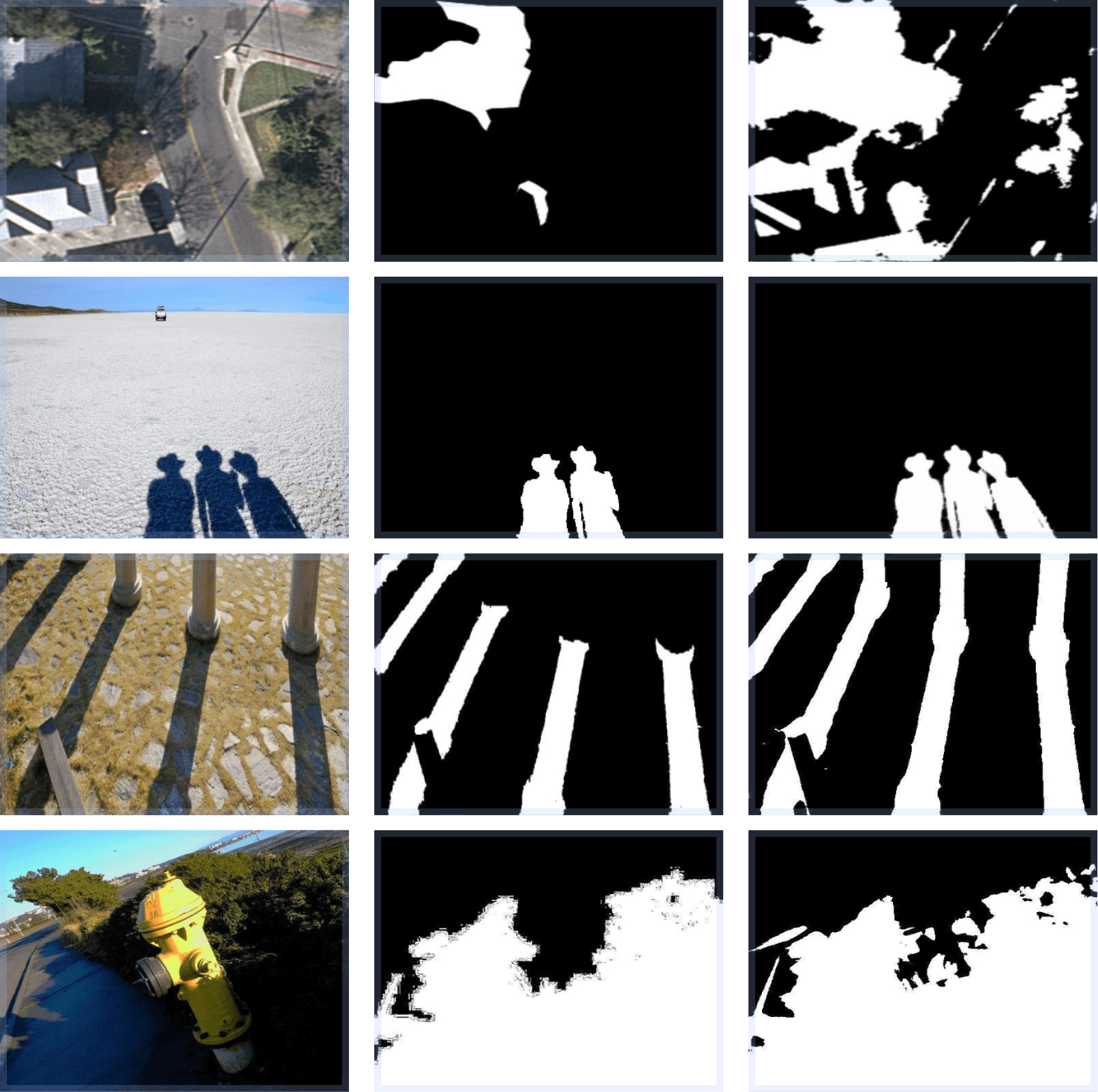}
	\end{minipage}
	\begin{minipage}[t]{0.32\linewidth}
        \centerline{\footnotesize \hspace*{2mm} (a) Training images}\vspace*{-1mm}
		\centerline{\footnotesize \hspace*{2mm} in existing datasets}
	\end{minipage}
	\begin{minipage}[t]{0.32\linewidth}
		\centerline{\footnotesize \hspace*{1mm} (b) Original labels}
		\centerline{\footnotesize \hspace*{1mm} in existing datasets}
	\end{minipage}
	\begin{minipage}[t]{0.32\linewidth}
		\centerline{\footnotesize \hspace*{-1mm} (c) Our refined}
		\centerline{\footnotesize \hspace*{-1mm} labels}
	\end{minipage}
	\caption{Column (b): Labels in existing datasets may not be accurate.
Column (c): Our automatically-refined annotations.
Note that row 1 is from UCF~\cite{zhu2010learning} and rows 2-4 are from SBU~\cite{vicente2016large}.}
	\label{fig:intro}
\end{figure}

Detecting shadows is very challenging,
since shadows have no specific shapes, colors, or textures, and their intensity may just be slightly lower than the surroundings.
Thanks to the advances in deep learning, many works~\cite{zhu2022single, zhu2021mitigating, chen2020multi, zheng2019distraction} have been developed and they show 
great progress
in detecting shadows. They mostly propose new and delicate network architectures and train the network directly on shadow datasets with labeled shadow regions.

However, it can be observed that labels in existing datasets~\cite{zhu2010learning,hou2019large} may not be accurate.
For example, as Fig.~\ref{fig:intro} shows, the training samples may lack details (row 1);
some shadows could be incomplete (row 2);
some self shadows may be missed (row 3); and the annotations could be rough (row 4).
Fundamentally, there are two main reasons.
First, 
the annotations of the SBU~\cite{vicente2016large} dataset are generated from a manually lazy-labeled dataset using an LSSVM-based method, so the resulting 
annotations can be noisy, and some detailed and background shadows may be ignored.
%
Second, the perception of shadow can be subjective, especially for self shadow and soft shadow. Since datasets are typically prepared by several human annotators, shadow and non-shadow regions could be labeled inconsistently in the same dataset. 
Hence, existing methods trained on such noisy datasets could be easily biased by the noisy labels, which hinder them to achieve better performance.

\begin{figure}[t]
\centering
\includegraphics[width=0.95\linewidth]{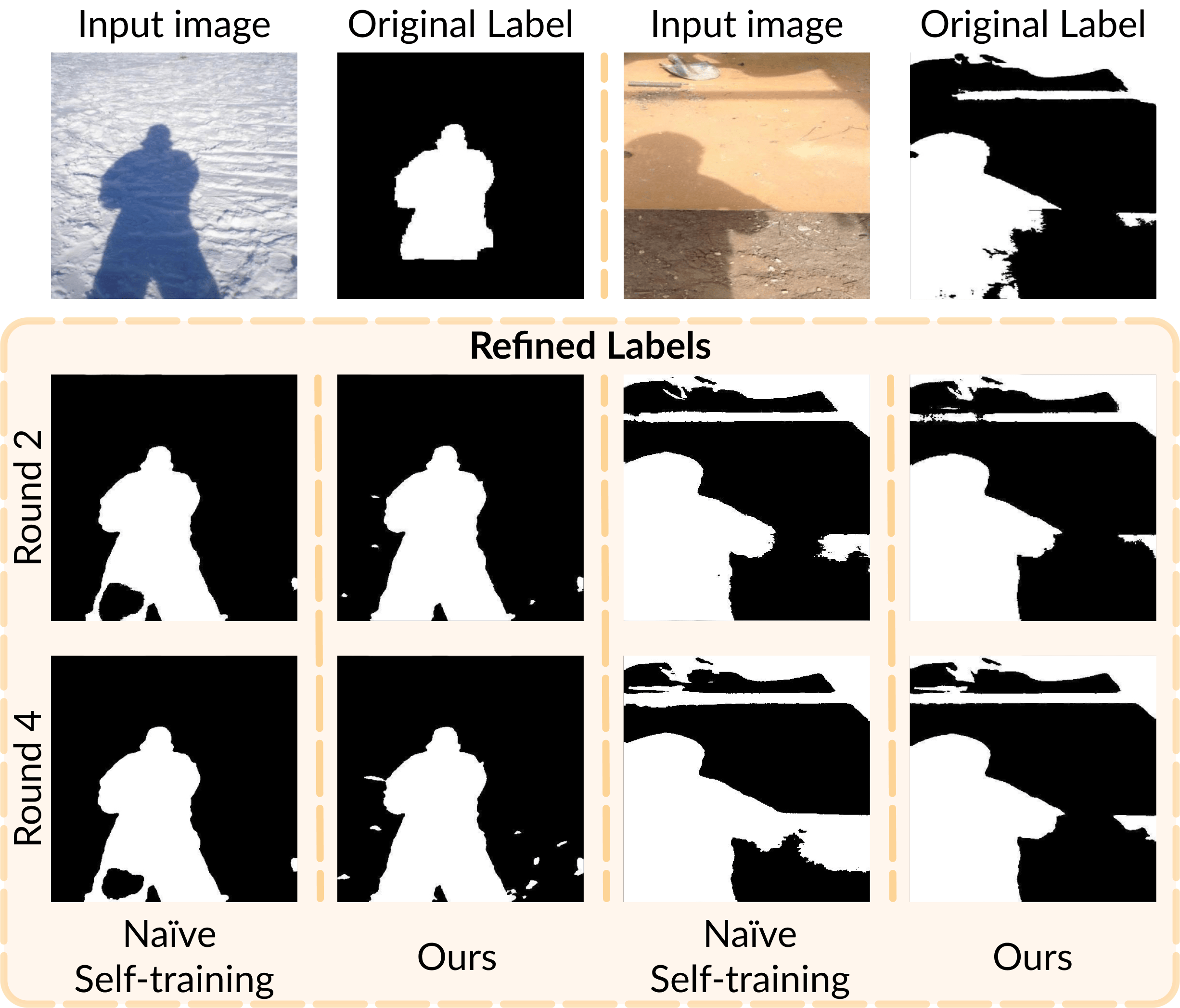}
\caption{Comparison of label quality produced by a na\"ive self-training framework and our SILT framework. Our SILT framework generates more accurate labels with finer details.
}
\label{fig:problem}
\end{figure}

To learn from data with noisy labels, it is intuitive to adopt a self-training framework~\cite{xie2020self},~\ie,
train a network on noisy data; use the trained network to relabel the training data; and repeat these two steps 
alternately to 
refine the labels.
%
%
Yet, directly adopting self-training to detect shadows may not work well for the following reasons:
\begin{itemize}
    \item Overfitting the training data. Considering the limited size of existing datasets, the trained network is prone to remember every single sample.
    So, if we directly use it to relabel the training data, the network may be too conservative to 
    try large refinements on the noisy labels.
    As Fig.~\ref{fig:problem} column 1 shows, the network could fail to label some obvious shadows due to overfitting.
    %
    \item Error accumulation.
    Existing methods often falsely predict dark objects as shadows.
    So, the iterative self-training process may easily accumulate such prediction errors and wrongly encourage the network to label all dark objects as shadows; see Fig.~\ref{fig:problem} column 3.
\end{itemize}



In this paper, we present the Shadow-aware Iterative Label Tuning (SILT) framework by formulating the following strategies to address the above challenges.
Firstly, we adopt a shadow-aware data augmentation strategy that involves shadow counterfeiting and incorporating dark regions and noise into input images. This enhances the ability of the deep network to recognize non-shadow regions by teaching it to identify these regions amidst the added noise.
Secondly, we propose a global-local fusion approach that involves splitting the input image into multiple patches and using the network to predict masks for each patch and the entire image. This approach helps alleviate overfitting. We then perform shadow-aware filtering that considers both the image brightness information and the previous shadow masks to select accurate shadow masks while filtering out inaccurate predictions.
Thirdly, we collect a set of zero-labeled non-shadow images with dark objects to train the network to better identify non-shadow regions.
Using the above techniques, we can effectively train a simple U-Net~\cite{ronneberger2015u} in SILT to iteratively refine the noisy labels in the original datasets. Examples demonstrating the effectiveness of the proposed approach are shown in Fig~\ref{fig:intro} (c).

For quantitative evaluation,
we relabel the test set of SBU~\cite{vicente2016large} to obtain high-quality and accurate shadow masks, due to the existing issue of noisy labels. 
With this carefully relabeled test set, we conduct various experiments and demonstrate the superiority of our approach in producing more precise shadow detection results compared to existing state-of-the-art methods. 
Furthermore, we find that our refined training set can significantly improve the performance of these state-of-the-art methods.
The code, pretrained model, and dataset are publicly available at \url{https://github.com/Cralence/SILT}.


\begin{figure*}[t]
\centering
\includegraphics[width=\textwidth]{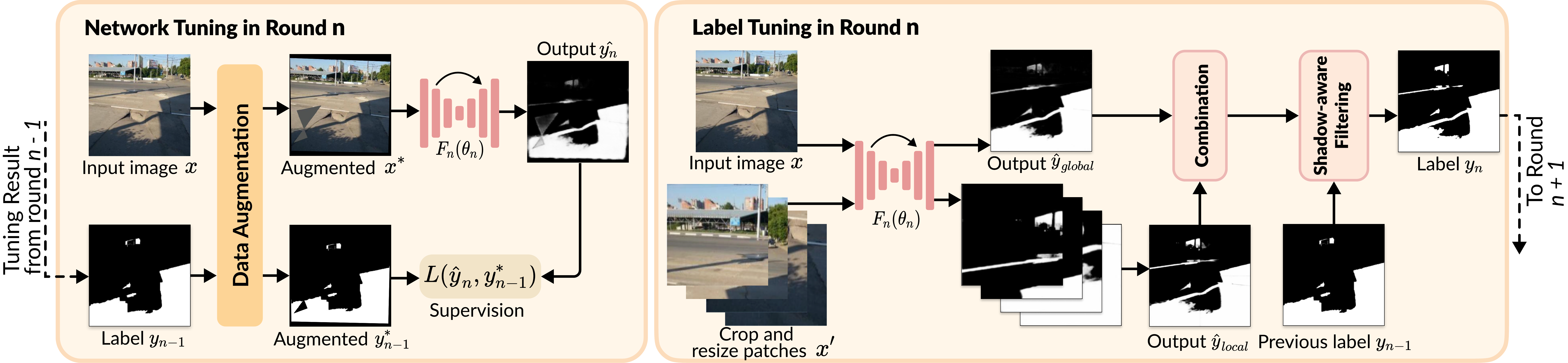}
\caption{The architecture of Shadow-aware Iterative Label Tuning (SILT) for learning to detect shadows from data with noisy labels.
}
	\label{Overview}
\end{figure*}

\section{Related Works}
\label{sec:relatedworks}

\paragraph{Shadow detection.} Single-image shadow detection has been studied for a long time. Earlier methods detect shadows mainly based on physical features such as geometrical properties~\cite{salvador2004cast, panagopoulos2011illumination}, spectrum ratios~\cite{tian2016new}, color~\cite{lalonde2010detecting, guo2011single, vicente2015leave, vicente2018leave}, texture~\cite{zhu2010learning, guo2011single, vicente2015leave, vicente2018leave}, edge~\cite{lalonde2010detecting, zhu2010learning, huang2011characterizes}, etc. 
However, due to the physical nature of the shadow, it is hard to quantify shadow by a few situation-invariant features. 
Therefore, these methods work well mainly on simple cases and are less effective on complex real-world shadows. 

With deep convolutional neural networks (CNN), features can be automatically learned instead of hand-crafted.
Since then, the performance of shadow detection improves significantly.
Khan et al.~\cite{khan2014automatic} first use a CNN followed by a conditional random field 
to achieve pixel-wise shadow detection.
After that, many architectures~\cite{shen2015shadow, vicente2016large, nguyen2017shadow, Hu_2018_CVPR, hu2019direction, wang2018stacked, le2018a+d, zhu2018bidirectional, zheng2019distraction, ding2019argan, chen2020multi, zhu2021mitigating, hu2021revisiting, zhu2022single, fang2021robust, zhou2022shadow, wu2022light, jie2022rmlanet, vicente2016noisy,Wang_2020_CVPR,Wang_2021_CVPR,Wang_2022_TPAMI} are proposed to adopt deep neural networks for shadow detection.
Besides, various high-level shadow features are designed,~\eg, direction-aware spatial context~\cite{Hu_2018_CVPR, hu2019direction}, distraction-aware~\cite{zheng2019distraction}, shadow edges~\cite{shen2015shadow, chen2020multi}, shadow count~\cite{chen2020multi}, etc.
Further, several datasets are built,~\eg, SBU~\cite{vicente2016large} and ISTD~\cite{wang2018stacked}.
%
Trained on these data, deep neural networks can learn to detect shadows in more situations.

However, prior works mostly do not consider the noisy label problem.
One relevant work~\cite{vicente2016large} is an LSSVM-based noisy label recovery method, which adopts image clustering to recover the annotations from a lazily labeled dataset.
However, the SBU dataset refined by this method still contains many noises.
%
In this work, we reformulate shadow detection as a noisy label problem by explicitly considering the noisy labels in the training data and present a new shadow-aware iterative label tuning framework to learn to detect shadows from noisy labels.


\paragraph{Learning with noisy labels.}
Existing works for the classification task have widely studied the noisy label issue.
Various aspects have been exploited,~\eg, network architectures~\cite{chen2015webly, xiao2015learning, goldberger2016training, han2018masking}, losses~\cite{ghosh2017robust, lyu2019curriculum, wang2019symmetric, zhang2018generalized, zhou2021asymmetric, zhou2021learning}, regularizations~\cite{menon2019can, tanno2019learning, xia2020robust},~\etc
%
On the other hand, some methods aim to first detect~\cite{ren2018learning, zhou2020robust} noises then re-weight~\cite{patrini2017making} or correct~\cite{tanaka2018joint, vahdat2017toward, veit2017learning, yi2019probabilistic, han2019deep} the detected noises. 
%
To perform image segmentation with 
noisy 
labels,
%
some methods improve the robustness of the network to noise by
designing meta-structures~\cite{luo2022deep} or leveraging reliability of annotators~\cite{zhang2020disentangling} and noisy gradient~\cite{min2019two}. 
Others correct errors by exploiting early-learning phase~\cite{liu2022adaptive}, using local visual cues~\cite{shu2019lvc} and formulating spatial label smoothing regularization~\cite{zhang2020characterizing}. 

. 


\vspace{-2.5mm}
\paragraph{Self-training.}
Self-training methods~\cite{yarowsky1995unsupervised, nigam2000analyzing, lee2013pseudo} typically first train a teacher network on a small labeled dataset.
Then, we can use it to generate pseudo-labels for a large unlabeled dataset and take the pseudo-labels to train a student network; by iterate this process, the quality of the pseudo-labels can be gradually improved.
%
Recently, self-training methods achieve state-of-the-art performance in tasks like image classification~\cite{li2019learning, xie2020self} and segmentation~\cite{mendel2020semi, ouali2020semi, li2019supervised}.

Our method differs from previous works in self-training and noisy labels in two aspects.
First, \emph{we only have small but noisy datasets}, while 
previous self-training methods typically adopt a huge unlabeled or badly-labeled dataset~\cite{lee2013pseudo, xie2020self, tanaka2018joint, han2019deep, xiao2015learning}, sometimes together with a small and clean dataset~\cite{lee2013pseudo, xie2020self}. Thus, we propose global-local fusion in label tuning to mitigate the over-fitting issue.
%
Second, \emph{in shadow detection, errors appear more with a common pattern}, compared with general image classification and segmentation, so it is hard to tackle the error accumulation issue.  Thus, we propose shadow counterfeiting in network tuning to enhance the discriminating ability of the deep network to recognize non-shadow regions and shadow-aware filtering in label tuning to select accurate shadow masks while filtering out inaccurate predictions.

\section{Methodology}
\label{sec:method}

Fig.~\ref{Overview} shows the overall architecture of the proposed Shadow-aware Iterative Label Tuning (SILT) framework, which automatically tunes the labels in the noisy dataset in a shadow-aware manner.  
%
For each round $n \in \{1, 2, ..., N\}$, where $N$ is the total number of rounds, 
we denote the input images as $X$, and its corresponding shadow masks as $Y_{n-1}$.
When $n = 1$, the corresponding $Y_0$ is the original noisy label. 
%
%
%
%
There are two stages in each round of  SILT: \textit{Network Tuning} and \textit{Label Tuning}.
In network tuning, we train a shadow detection network supervised by  previously tuned labels $Y_{n-1}$, during which we introduce strong data augmentation with shadow counterfeiting; see details in Sec.~\ref{subsec:network}. 
In label tuning, we tune the shadow masks using the previous freeze-weight network with 
global-local fusion
and shadow-aware filtering strategies;
see details in Sec.~\ref{subsec:tuning}. By iteratively performing these two stages, the labels gradually become more accurate and contain more fine details.



\subsection{Network Tuning with Shadow Counterfeiting}~\label{subsec:network}
%
%
%
%

Dark objects are easily misclassified as shadow regions and the errors could be accumulated in the self-training frameworks~\cite{xie2020self, han2019deep}.
In order to let the network learn to distinguish the shadow and dark non-shadow objects, we introduce
the shadow counterfeiting in network tuning.
This strategy is implemented by two data augmentations, namely Distraction and
RandomNoise, which add different types of dark regions (counterfeited shadows) to train the network to recognize the dark regions as non-shadows.

\begin{figure}[t]
    \centering
\includegraphics[width=0.97\linewidth]{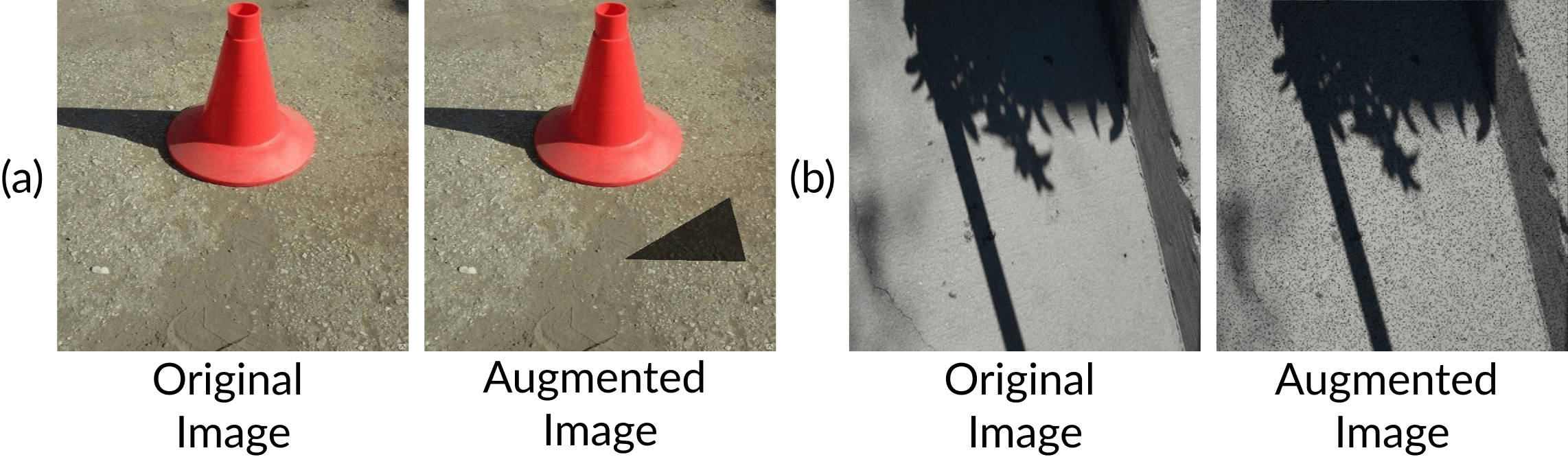}
\caption{An illustration of shadow counterfeiting.}
\label{fig:data augmentation}
\end{figure}

Specifically, in Distraction (Fig.~\ref{fig:data augmentation} (a)), we randomly choose an area in the training image and fill it with dark color. 
The area is a randomly generated polygon, and we multiply the value of the pixels in that polygon by a factor of 0.3.
In RandomNoise (Fig.~\ref{fig:data augmentation} (b)), we randomly add black points in the training images. 
%
We label both augmentations' newly added regions/pixels as {\bf non-shadow regions}. These two augmentations introduce "counterfeited" shadows with various sizes and numbers to the training data, 
thereby improving its ability to distinguish between shadows and dark non-shadows. To mitigate overfitting, we add other common data augmentations following previous works~\cite{yuan2021simple,cubuk2019randaugment}.

\subsection{Label Tuning}~\label{subsec:tuning}

%
In label tuning, we aim to employ the network to correct wrong labels and add missing labels in the training data.
However, the network may easily overfit the wrong labels since the shadow detection datasets~\cite{zhu2010learning,vicente2016large} are relatively small compared with other datasets used for general computer vision tasks, \eg, ImageNet~\cite{deng2009imagenet} for image classification and COCO~\cite{lin2014microsoft} for object detection. 
To solve the above issues, we present the label tuning strategy, which first adopts global-local fusion to enhance details as well as alleviate overfitting and then uses shadow-aware filtering to filter out inaccurate and vague predictions.
%

\subsubsection{Global-Local Fusion}
As shown in the right part of Fig.~\ref{Overview},
unlike the previous works~\cite{li2019learning, xie2020self} that use the network to relabel their training data directly, we first split the input image into four parts, resize them to the size of the original input image, and adopt the network to predict the shadow mask for each part. Lastly, we obtain a unified output $\hat{y}_{local}$, referred to as local-view prediction, by combining different local parts.

\begin{figure}[tp]
    \centering
    \includegraphics[width=0.93\linewidth]{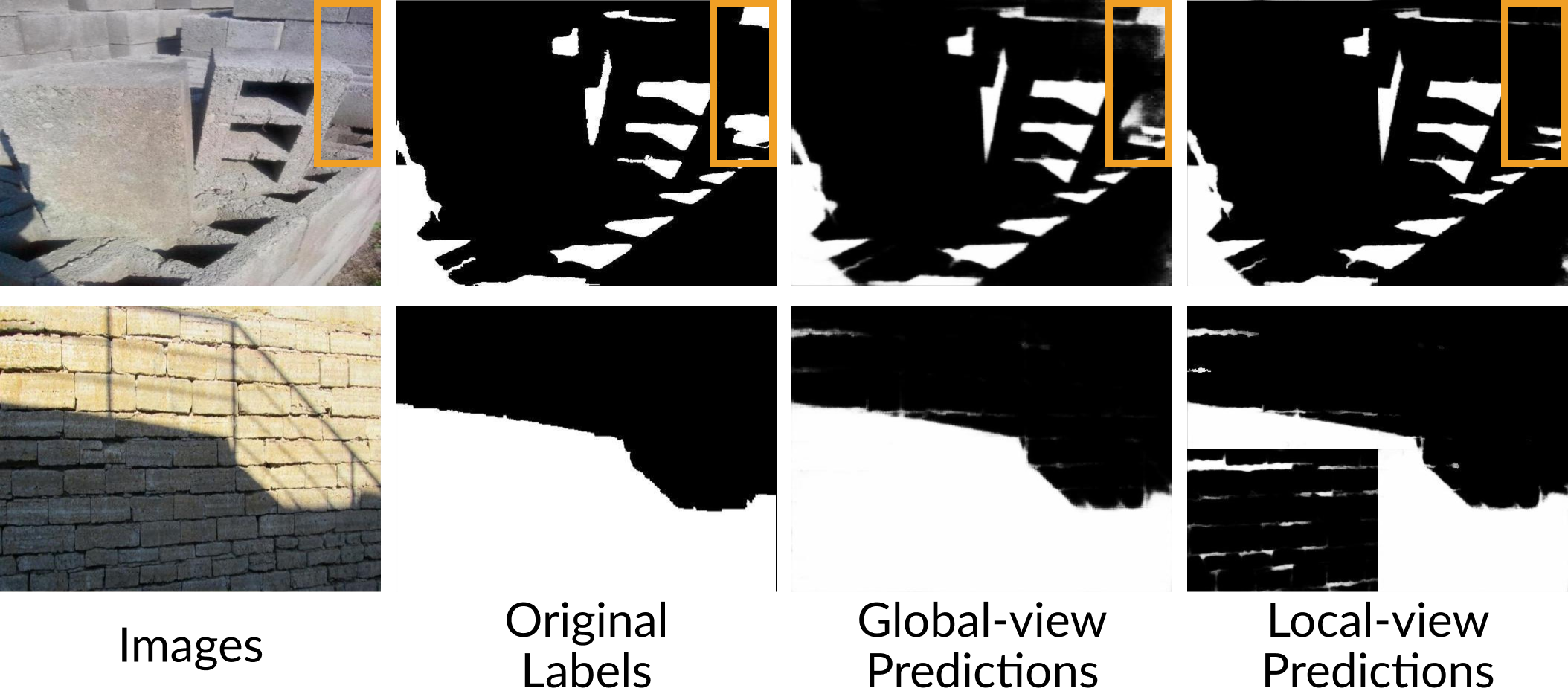}
    \caption{Visualizations of global-view and local-view prediction.}
    \label{fig:global_local_view}
\end{figure}

The benefits of such operations are two-fold. First, inspired by the concept of multi-scale inference~\cite{najibi2019autofocus}, a higher resolution input image helps the network to predict more detailed shadow masks.  More importantly, taking crop-and-resized patches as input avoids directly using the raw training data, which mitigates overfitting and encourages noisy label correction, as depicted in the first row of Fig.~\ref{fig:global_local_view}.

Besides the local-view prediction, we further take the whole image as input to generate the global-view prediction $\hat{y}_{global}$,
which helps to distinguish the large shadow region by considering the global image context, as shown in the second row in Fig.~\ref{fig:global_local_view}.
After obtaining the predictions from the global view and the local view, 
we combine them into the final prediction $\hat{y_n}$ by  
\begin{equation}
\hat{y_n} =\begin{cases}
\text{max}(\hat{y}_{global},\hat{y}_{local}), & \hat{y}_{global}>R_{filt}\\
\hat{y}_{global}, & \hat{y}_{global}\leqslant R_{filt} \\
\end{cases} \ , 
\end{equation}
where $R_{filt}$ is the threshold to ensure the priority of $\hat{y}_{global}$, since the global-view result is more reliable by considering the global image context. We empirically set the $R_{filt}$ as $0.1$ during the experiments.

\if{
\begin{figure}[b]
    \centering
    \includegraphics[width=0.7\linewidth]{figs/Group 154.png}
    \caption{A scatter plot showing the brightness and confidence relation in shadow prediction. Each dot represents a pixel in the predicted shadow masks.}
    \label{fig:scatter_plot}
\end{figure}
}\fi

\subsubsection{Shadow-aware Filtering}
The prediction $\hat{y_n}$ contains the continuous values, and some ambiguous regions, \eg dark non-shadow objects, usually have low confidence.
%
If we keep these ambiguous regions, the confidence value will accumulate stage by stage and finally mislead the network to recognize these regions as shadow regions with high confidence.
%
To avoid this situation, we pass the prediction $\hat{y_n}$ through a threshold map to generate a binary mask.

We observe that the network often has lower confidence in predicting shadows in bright regions while has higher confidence in dark regions.  But the dark regions are easily to be mislabeled.
Therefore, we prefer to adopt a lower threshold for bright regions, 
 thus making the network easier to refine the wrong labels in those regions. 
Meanwhile, we prefer a higher threshold for dark regions, which helps to filter out the wrong labels of dark non-shadow objects predicted by the network. 
For this purpose, we construct a shadow-aware filter map to perform binarization with different thresholds according to the image brightness.


Specifically, to obtain the illumination intensity, we transform the RGB image into $YC_bC_r$ color space~\cite{jin2010towards} and take the first channel as brightness map $Y$.
Then, we construct a shadow-aware filter map $F$, which gives a smaller threshold when the brightness is high and vice versa.
In detail, we set $R_{min}$ and $R_{max}$ as two hyper-parameters 
denoting the minimum and maximum value of the threshold map, respectively,
and for each pixel $i$, the corresponding threshold $F_i$ is defined as:
\begin{equation}
\label{eq:2}
    F_i=\frac{R_{max}Y_{max} - R_{min}Y_{min}}{Y_{max}-Y_{min}} - \frac{R_{max}-R_{min}}{Y_{max}-Y_{min}} \cdot Y_i \ ,
\end{equation}
where $Y_{max}$ and $Y_{min}$ denote the  maximum and minimum brightness value of the image, respectively, and $Y_i$ is the brightness at the pixel $i$.
Note that $F_i$ is defined as $R_{max}$ when $Y_i$ takes $Y_{min}$ and as $R_{min}$ when $Y_i$ takes $Y_{max}$.
During the experiments, we empirically set $R_{min}$ as 0.5 and $R_{max}$ as 0.6.
%
Through the above formulation, we adopt a higher threshold to make a strict selection in dark regions and use a lower threshold to avoid filtering out correct masks in bright regions. 

After we obtain the newly generated shadow mask $\hat{y_n}$ , we compare it with the previous shadow mask $y_{n-1}$ and use it to replace the previous shadow mask when $\hat{y_n}$ contains at least $R_{corr}$ percentage of the previous shadow mask. Otherwise, we will keep the previous shadow mask as the final result. 
This is because the major part of the original/previous shadow mask is usually reliable, especially in the larger regions.
We empirically set $R_{corr}$ as 0.95 in the following experiments.

\section{Experiments}
\label{sec:experiments}

\begin{table*}[htbp]
\centering
\caption{Quantitative Comparison of our method with recent state-of-the-art methods on three benchmark datasets. (a) and (b) are evaluated on our relabeled SBU test set; (c) is evaluated on original ISTD test set.} 
\label{tabel:comp1}
\renewcommand\tabcolsep{8.0pt}
\resizebox{\textwidth}{!}{
\begin{tabular}{c|c|c|ccc|ccc|ccc} 
\toprule
\multicolumn{2}{c|}{Training set}                                                                                                                                             & Param.(M)  & \multicolumn{3}{c|}{SBU (a)~} & \multicolumn{3}{c|}{UCF (b)~} & \multicolumn{3}{c}{ISTD (c)~} \\ 
\hline
Method & Year &  & BER$\downarrow$ & $\text{BER}_S$  & $\text{BER}_{NS}$  & BER$\downarrow$   & $\text{BER}_S$   & $\text{BER}_{NS}$   & BER$\downarrow$   & $\text{BER}_S$    & $\text{BER}_{NS}$  \\ 
\hline \hline
\multicolumn{1}{l|}{\begin{tabular}[c]{@{}l@{}}BDRAR~\\ DSC~\\ DSD~\\ MTMT~\\FSD~\\ FDRnet~\\SDCM~\end{tabular}} & \begin{tabular}[c]{@{}c@{}}ECCV2018\\ CVPR2018\\ CVPR2019\\ ICCV2021\\ TIP2021\\ ICCV2021\\ ACM MM22\end{tabular}    & \begin{tabular}[c]{@{}c@{}}42.46 \\ 79.03 \\ 58.16 \\ 44.13 \\ 4.40 \\ 10.77 \\ 10.95\end{tabular} & \begin{tabular}[c]{@{}c@{}}6.49\\ 8.08\\ 5.60\\ 7.41\\10.87\\ 5.93\\ 5.71\end{tabular}       & \begin{tabular}[c]{@{}c@{}}9.68\\ 12.25\\ 7.86\\ 13.68\\20.39\\ 10.93\\9.07\end{tabular}        & \begin{tabular}[c]{@{}c@{}}3.29\\ 3.91\\ 3.34\\ 0.97\\1.34\\ 1.71\\ 2.36\end{tabular} & \begin{tabular}[c]{@{}c@{}}11.48\\14.15\\ -\\ -\\17.66\\12.91\\ 11.45\end{tabular}        & \begin{tabular}[c]{@{}c@{}}18.81\\24.86\\ -\\ -\\32.38\\22.50\\ 20.69\end{tabular}            & \begin{tabular}[c]{@{}c@{}}4.15\\3.44\\ -\\-\\2.93\\3.32\\ 2.22\end{tabular}    & \begin{tabular}[c]{@{}c@{}}2.69\\3.42\\2.17\\1.72\\2.68\\1.55\\1.41\end{tabular} & \begin{tabular}[c]{@{}c@{}}0.50\\3.85\\1.36\\1.36\\3.69\\1.22\\1.19\end{tabular} & \begin{tabular}[c]{@{}c@{}}4.87\\3.00\\2.98\\2.08\\1.66\\1.88\\1.69\end{tabular}  \\ 
\hline \hline
\multicolumn{2}{l|}{\begin{tabular}[c]{@{}l@{}}ours (EfficientNet-B3)\\ ours (ConvNeXt-B)\\ ours (ResNeXt-101)\\ ours (EfficientNet-B7)\\ ours (PVT v2-B3)\\ ours (PVT v2-B5)\end{tabular}}   & \begin{tabular}[c]{@{}c@{}}12.18 \\ 100.68 \\ 90.50 \\67.80 \\ 49.42 \\ 86.14\end{tabular}  & \begin{tabular}[c]{@{}c@{}}5.23\\5.11\\5.08\\ 4.62\\ 4.36 \\\textbf{4.19}\end{tabular} & \begin{tabular}[c]{@{}c@{}}6.22\\ 7.07\\4.86\\ \textbf{4.24} \\5.29 \\ 4.28\end{tabular} & \begin{tabular}[c]{@{}c@{}}4.23\\3.15\\5.30\\ 4.90 \\ 3.43\\ 4.09\end{tabular} & \begin{tabular}[c]{@{}c@{}}9.18\\8.62\\9.27\\7.97\\ 7.25 \\ \textbf{7.23} \end{tabular} & \begin{tabular}[c]{@{}c@{}}12.32\\11.54\\ 12.71\\9.41 \\ \textbf{7.39}\\ 7.78\end{tabular} & \begin{tabular}[c]{@{}c@{}}6.04\\5.70\\5.82\\6.54 \\7.12 \\ 6.69\end{tabular} & \begin{tabular}[c]{@{}c@{}}2.00\\1.15\\1.53\\1.46\\\textbf{1.11}\\1.16\end{tabular}                & \begin{tabular}[c]{@{}c@{}}1.62\\0.82\\1.20\\1.01\\0.79\\0.85\end{tabular}                & \begin{tabular}[c]{@{}c@{}}2.37\\1.48\\1.86\\1.90\\\textbf{1.44}\\1.47\end{tabular}                 \\
\bottomrule
\end{tabular}}
\end{table*}

\begin{table*}[htbp]
\centering
\caption{Comparison results of the recent state-of-the-art methods trained on the original dataset and our refined dataset. Evaluations are done on our relabeled SBU test set.}
\vspace{1mm}
\renewcommand\tabcolsep{10.0pt}
\resizebox{0.98\textwidth}{!}{%
    \begin{tabular}{llc|c|ccc|cc}
    \toprule
    \multicolumn{3}{l|}{} & Trained on original dataset & \multicolumn{3}{c|}{Trained on our refined dataset}  & \multicolumn{2}{c}{\% Reduction} \\ \hline
    \multicolumn{2}{c|}{Method}  & Year & BER$\downarrow$  & BER$\downarrow$  & $\text{BER}_S$$\downarrow$  & $\text{BER}_{NS}$$\downarrow$   & BER   & $\text{BER}_S$  \\ \hline\hline
    \multicolumn{1}{l|}{SBU~\cite{vicente2016large}} & \multicolumn{1}{l|}{\begin{tabular}[c]{@{}l@{}}
    BDRAR~\cite{zhu2018bidirectional}\\ DSC~\cite{Hu_2018_CVPR}\\ 
    FSD~\cite{hu2021revisiting}\\ FDRnet~\cite{zhu2021mitigating}\\ SDCM~\cite{zhu2022single}\end{tabular}} & \begin{tabular}[c]{@{}c@{}}ECCV2018\\ CVPR2018\\ TIP2021\\ ICCV2021\\ ACM MM22\end{tabular} & \begin{tabular}[c]{@{}c@{}}6.49\\8.08\\10.87\\   5.93\\ 5.71\end{tabular}     & \begin{tabular}[c]{@{}c@{}}5.35\\5.62\\6.13\\   5.48\\ 4.87\end{tabular}    & \begin{tabular}[c]{@{}c@{}}5.57\\5.38\\8.19\\   7.44\\ 4.32\end{tabular}     & \begin{tabular}[c]{@{}c@{}} 5.12\\5.86\\4.08\\  3.52\\ 5.41\end{tabular} & \begin{tabular}[c]{@{}c@{}}17.6\%\\30.4\%\\43.6\%\\   7.5\%\\ 14.4\%\end{tabular} & \begin{tabular}[c]{@{}c@{}}42.5\%\\56.1\%\\59.8\%\\   31.9\%\\ 55.7\%\end{tabular} \\ \hline\hline
    \multicolumn{1}{l|}{UCF~\cite{zhu2010learning}}  & \multicolumn{1}{l|}{\begin{tabular}[c]{@{}l@{}}
    BDRAR~\cite{zhu2018bidirectional}\\ DSC~\cite{Hu_2018_CVPR}\\ 
    FSD~\cite{hu2021revisiting}\\ FDRnet~\cite{zhu2021mitigating}\\ SDCM~\cite{zhu2022single}\end{tabular}} & \begin{tabular}[c]{@{}c@{}}ECCV2018\\ CVPR2018\\ TIP2021\\ ICCV2021\\ ACM MM22\end{tabular} & \begin{tabular}[c]{@{}c@{}}11.45\\14.15\\17.66\\   12.91\\ 11.45\end{tabular} & \begin{tabular}[c]{@{}c@{}} 9.59\\10.71\\13.39\\  9.12\\ 8.37\end{tabular} & \begin{tabular}[c]{@{}c@{}}9.55\\12.49\\21.21\\   11.97\\ 8.23\end{tabular} & \begin{tabular}[c]{@{}c@{}}9.64\\8.93\\5.57\\   6.27\\ 8.51\end{tabular} & \begin{tabular}[c]{@{}c@{}}16.2\%\\24.3\%\\24.2\%\\   29.4\%\\ 26.9\%\end{tabular} & \begin{tabular}[c]{@{}c@{}}49.2\%\\49.8\%\\34.5\%\\  72.1\%\\ 60.2\% \end{tabular} \\  \bottomrule
    \end{tabular} 
    }
\label{comp_performance_improvement}
\end{table*}

\subsection{Datasets and Evaluation Metrics}

We employ three mostly-used datasets. 
The first one is SBU~\cite{vicente2016large}, which contains 4089 training images and 638 testing images. The second is UCF Shadow Dataset~\cite{zhu2010learning}, which includes 360 images. The third is ISTD~\cite{wang2018stacked}, which has 1330 training images and 540 testing images. SBU and UCF contain more noisy labels, whereas ISTD is relatively cleaner. 
%
For evaluation metrics, we choose the most widely-used Balanced Error Rate (BER), Shadow Error Rate ($\text{BER}_{S}$), and Non-shadow Error Rate ($\text{BER}_{NS}$).

\subsubsection{Relabeled SBU test set}

Considering that the test sets in existing shadow datasets also contain wrong labels, we decide to relabel the test set of SBU~\cite{vicente2016large} to better evaluate the performance of various shadow detection networks.
%
Specifically, we hired three experts to do the relabeling work. 
Before relabeling, we showed them some examples to clarify the definitions and identifications of shadow, especially the self shadow. 
Then, they used Affinity Photo~\cite{Affinity} on the iPad with Apple Pencil to draw shadow labels image by image. 
Specifically, they were required to zoom in to label the details. 
It took on average five minutes to relabel each image. 
Finally, we cross-validated the refined labels from three experts and integrated them to obtain the final ground truth masks. 

\begin{figure}[tp]
    \centering
    \includegraphics[width=0.95\linewidth]{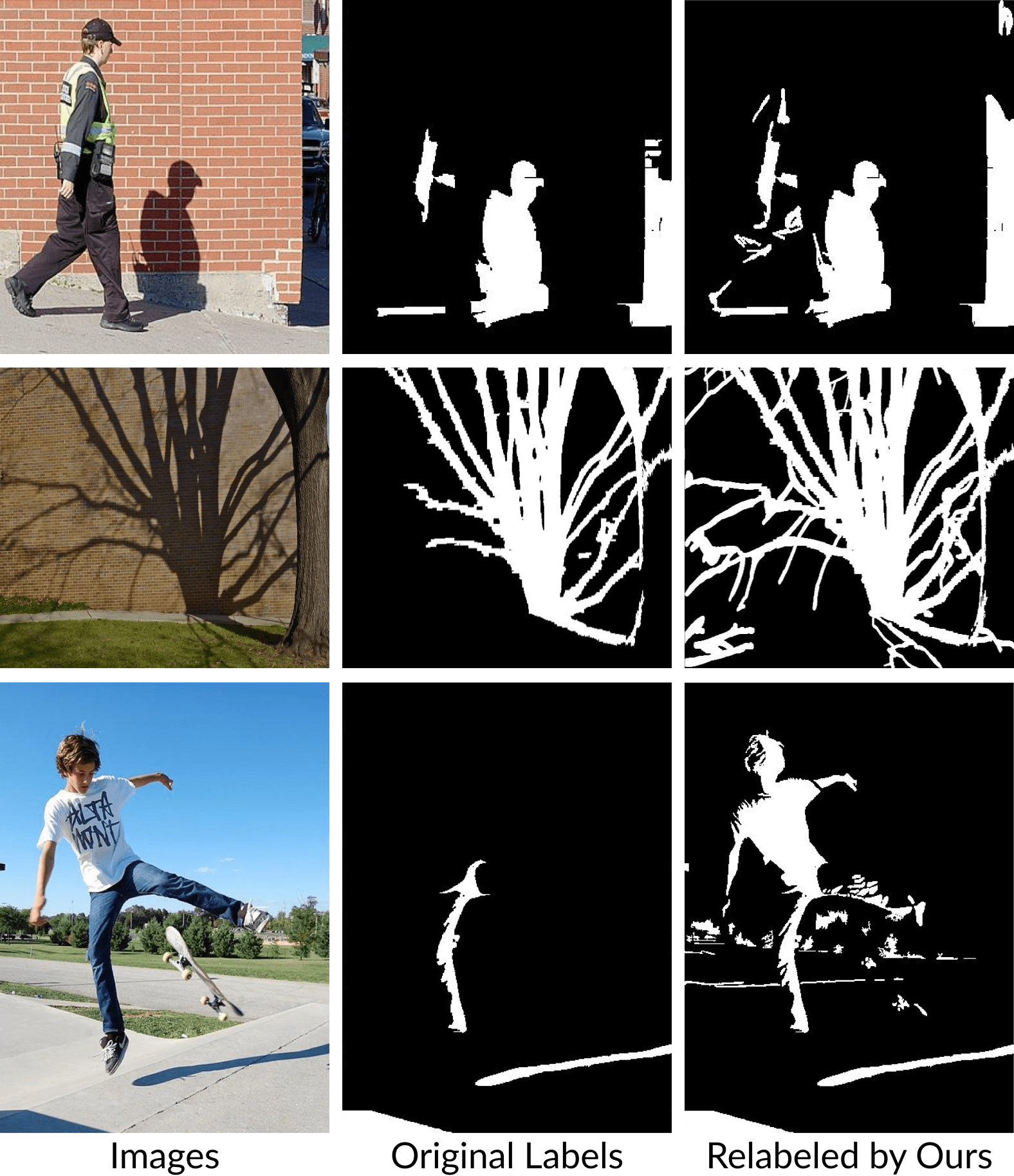}
    \caption{Comparison of our relabeled SBU test set and original SBU test set.}
    \label{fig:testset_comp}
\end{figure}

In total, our relabeled test set contains 638 test images and masks with fine details. In the original test set, $11.34\%$ of the mask pixels have been changed. Among them, $11.05\%$ are incomplete masks (new=1, old=0), while only $0.29\%$ are wrongly-labeled masks (new=0, old=1). Therefore, incomplete masks contribute the most to inaccurate annotations. 
Meanwhile, out of the 638 masks in our relabeled test set, 517 of them have modifications larger than $5\%$, and 374 of them have modifications larger than $20\%$. Figure~\ref{fig:testset_comp} shows comparison examples between our relabeled test set and the original SBU test set. Ours features more detailed shadow masks and exhibits more consistent labeling regarding self shadows and background shadows.

\begin{figure}[tp]
    \centering
    \includegraphics[width=\linewidth]{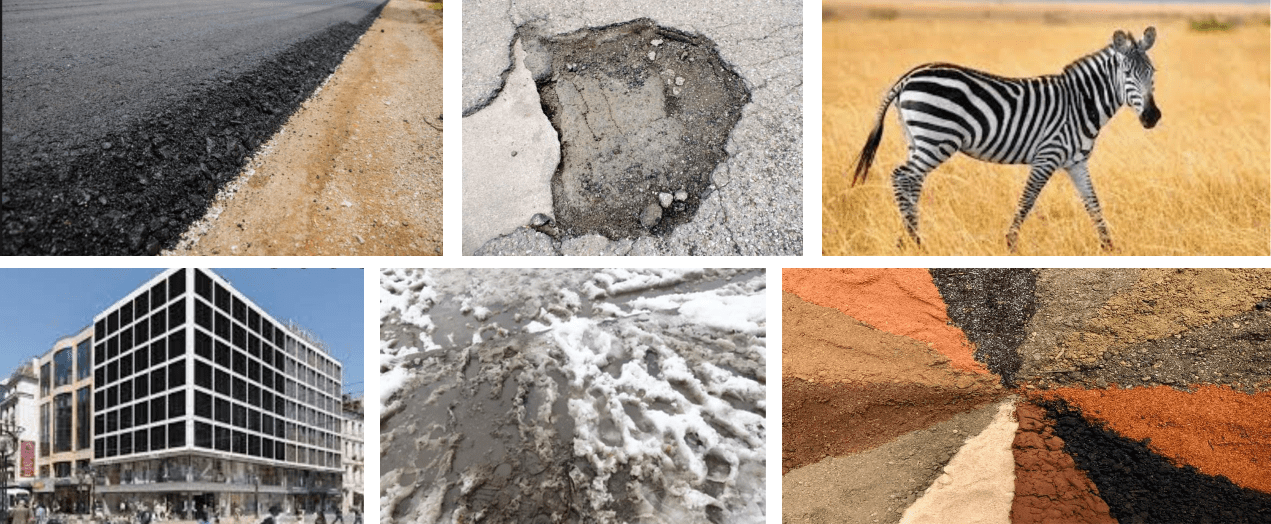}
    \caption{Examples of our additional dataset.}
    \label{fig:Additional dataset}
\end{figure}

\subsubsection{Additional Non-shadow Training Data}

We collect some images from the Internet to further help train the network to distinguish shadows and dark objects and alleviate error accumulation problem. 
Specifically, we search for some images that contain dark objects but no shadows, see Figure~\ref{fig:Additional dataset}.
Therefore, their shadow masks should be zeros, and no hand labeling is needed. 
Considering that too many training images with zeros as labels may deteriorate training, we only choose $89$ images in total as our additional training data, which is $2\%$ of the total number of the SBU. 
Note that we only use this additional training data when training on SBU~\cite{vicente2016large}, as the size of the training set of UCF~\cite{zhu2010learning} and ISTD~\cite{wang2018stacked} is very small.


\begin{figure*}[tpb]
    \centering
	\begin{minipage}[t]{\linewidth}
	    \centering
		\includegraphics[width=0.99\textwidth]{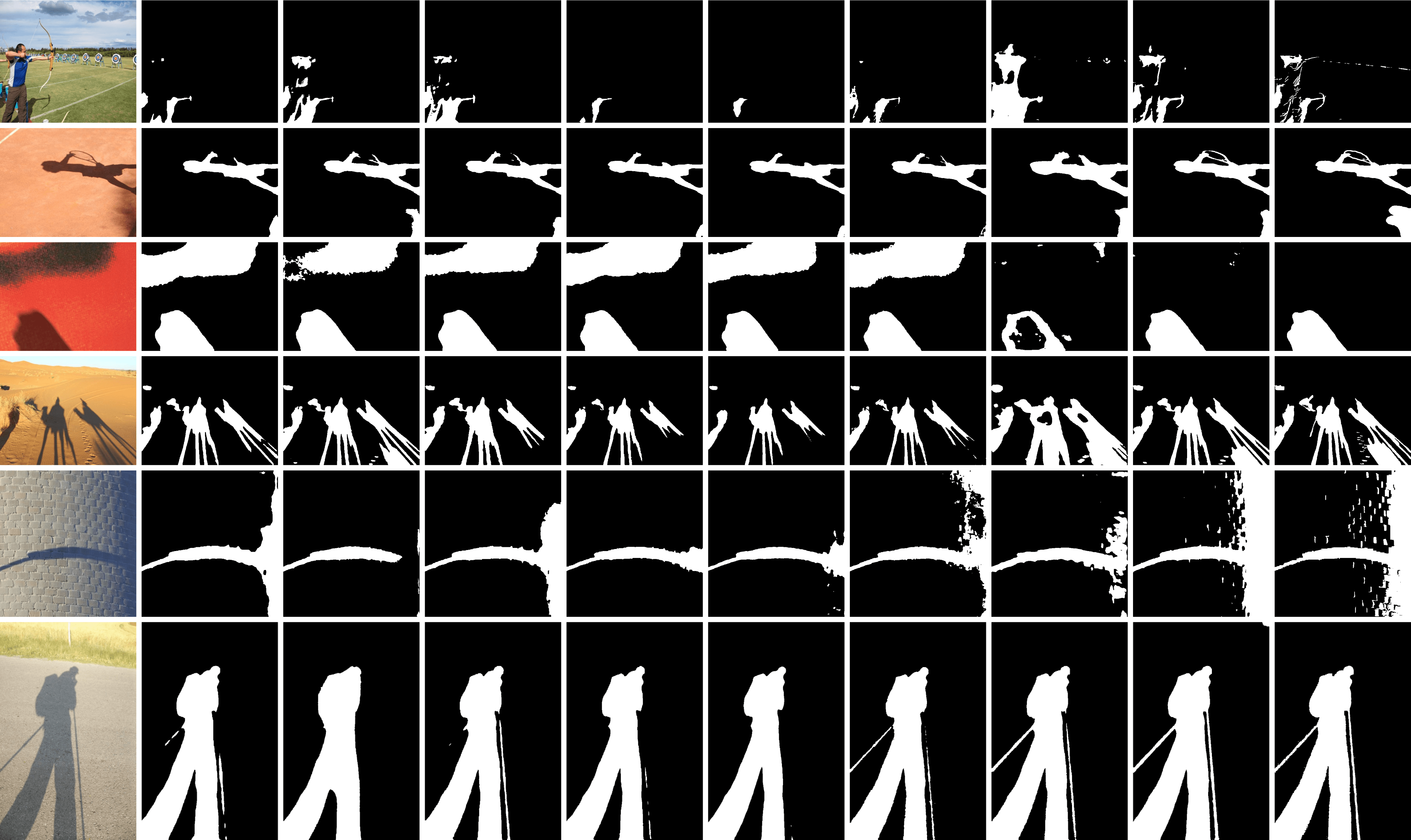}
		\vspace*{1mm}
	\end{minipage}
	\begin{minipage}[t]{0.096\textwidth}
		\centerline{\footnotesize (a) Input images}\vspace*{-1mm}
	\end{minipage}
	\begin{minipage}[t]{0.095\textwidth}
		\centerline{\footnotesize (b) BDRAR }\vspace*{-1mm}
        \centerline{\footnotesize \cite{zhu2018bidirectional} }
	\end{minipage}
	\begin{minipage}[t]{0.095\textwidth}
		\centerline{\footnotesize (c) DSC}\vspace*{-1mm}
          \centerline{\footnotesize \cite{Hu_2018_CVPR} }
	\end{minipage}
	\begin{minipage}[t]{0.095\textwidth}
		\centerline{\footnotesize (d) DSD}\vspace*{-1mm}
        \centerline{\footnotesize \cite{zheng2019distraction} }
	\end{minipage}
	\begin{minipage}[t]{0.095\textwidth}
		\centerline{\footnotesize (e) MTMT}\vspace*{-1mm}
        \centerline{\footnotesize \cite{chen2020multi} }
	\end{minipage}
	\begin{minipage}[t]{0.095\textwidth}
		\centerline{\footnotesize (f) FSD}\vspace*{-1mm}
        \centerline{\footnotesize \cite{hu2021revisiting} }
	\end{minipage}
	\begin{minipage}[t]{0.095\textwidth}
		\centerline{\footnotesize (g) FDRnet }\vspace*{-1mm}
        \centerline{\footnotesize \cite{zhu2021mitigating} }
	\end{minipage}
	\begin{minipage}[t]{0.095\textwidth}
		\centerline{\footnotesize (h) SDCM}\vspace*{-1mm}
        \centerline{\footnotesize \cite{zhu2022single} }
	\end{minipage}
	\begin{minipage}[t]{0.099\textwidth}
		\centerline{\footnotesize (i) Ours}\vspace*{-1mm}
	\end{minipage}
	\begin{minipage}[t]{0.095\textwidth}
		\centerline{\footnotesize (j) Ground truth}\vspace*{-1mm}
		\centerline{\footnotesize  }
	\end{minipage}
    \vspace{1mm}
    \caption{Qualitative comparison of our method with recent state-of-the-art methods.}
    \label{qualitative_comp}
\end{figure*}

\subsection{Experiment Details}
We adopt a simple {U-Net} structure~\cite{ronneberger2015u} as our shadow detection network. 
For a fair comparison, we use different encoder networks with similar parameter size as previous SOTA networks,~\ie, ResNeXt-101~\cite{xie2017aggregated}, ConvNeXt-B~\cite{liu2022convnet}, EfficientNet-B3~\cite{tan2019efficientnet}, EfficientNet-B7~\cite{tan2019efficientnet}, PVT v2-B3~\cite{wang2022pvt}, and PVT v2-B5~\cite{wang2022pvt}. 
Following previous works, we initialize the backbones with the weights pre-trained on ImageNet~\cite{deng2009imagenet}. 
We set the total number of self-training rounds $N$ as seven, and in each round, we train the network for 20 epochs with a batch size of six. 
We set the learning rate to $1\times10^{-4}$ for PVTs and $5\times10^{-4}$ for other networks, and use Adamax~\cite{adam} as the optimizer. We resize each image to $512\times512$ while training, refining, and inferring, 
and calculate the BERs on the original image size.
For models trained on SBU and UCF, we test the models on our
relabeled SBU test set. Besides, we employ the original ISTD test set to evaluate the models trained on ISTD, as the test set in ISTD is better labeled.
Note that we obtain the best result in round 6, 5, and 3 for models trained on SBU, UCF, and ISTD, respectively.

\subsection{Comparison with the State-of-the-art Methods}
We compare our training framework with seven previous SOTA networks, namely BDRAR~\cite{zhu2018bidirectional}, DSC~\cite{Hu_2018_CVPR,hu2019direction}, DSD~\cite{zheng2019distraction},  MTMT~\cite{chen2020multi}, FSD~\cite{hu2021revisiting}, FDRnet~\cite{zhu2021mitigating}, SDCM~\cite{zhu2022single}. 
We use their public pre-trained models on the SBU and ISTD, and re-trained models on the UCF for evaluation.

Table~\ref{tabel:comp1} shows the results.
We can see that our U-Net with various backbones outperforms all the previous SOTA networks by a large margin. 
On SBU, the PVTv2-based variant achieves the best performance, with a 25.2\% and 45.5\% decrease in BER and $\text{BER}_S$ compared with the best previous work. 
On UCF, our PVTv2-based variant achieves a 36.9\% and 60.7\% decrease in BER and $\text{BER}_S$ compared with the best previous work.
Note that we do not re-train MTMT and DSD, which require additional training data.
On ISTD, our PVTv2-based variant also reduces the BER and $\text{BER}_{NS}$ by 21.3\% and 13.3\%, respectively.



Fig.~\ref{qualitative_comp} shows the qualitative comparisons of our method with the recent state-of-the-art methods. From the results, we can see that our method is able to detect more fine shadow details with the help of the proposed SILT.


\if{
\begin{figure}[]
    \centering
    \includegraphics[width=0.95\linewidth]{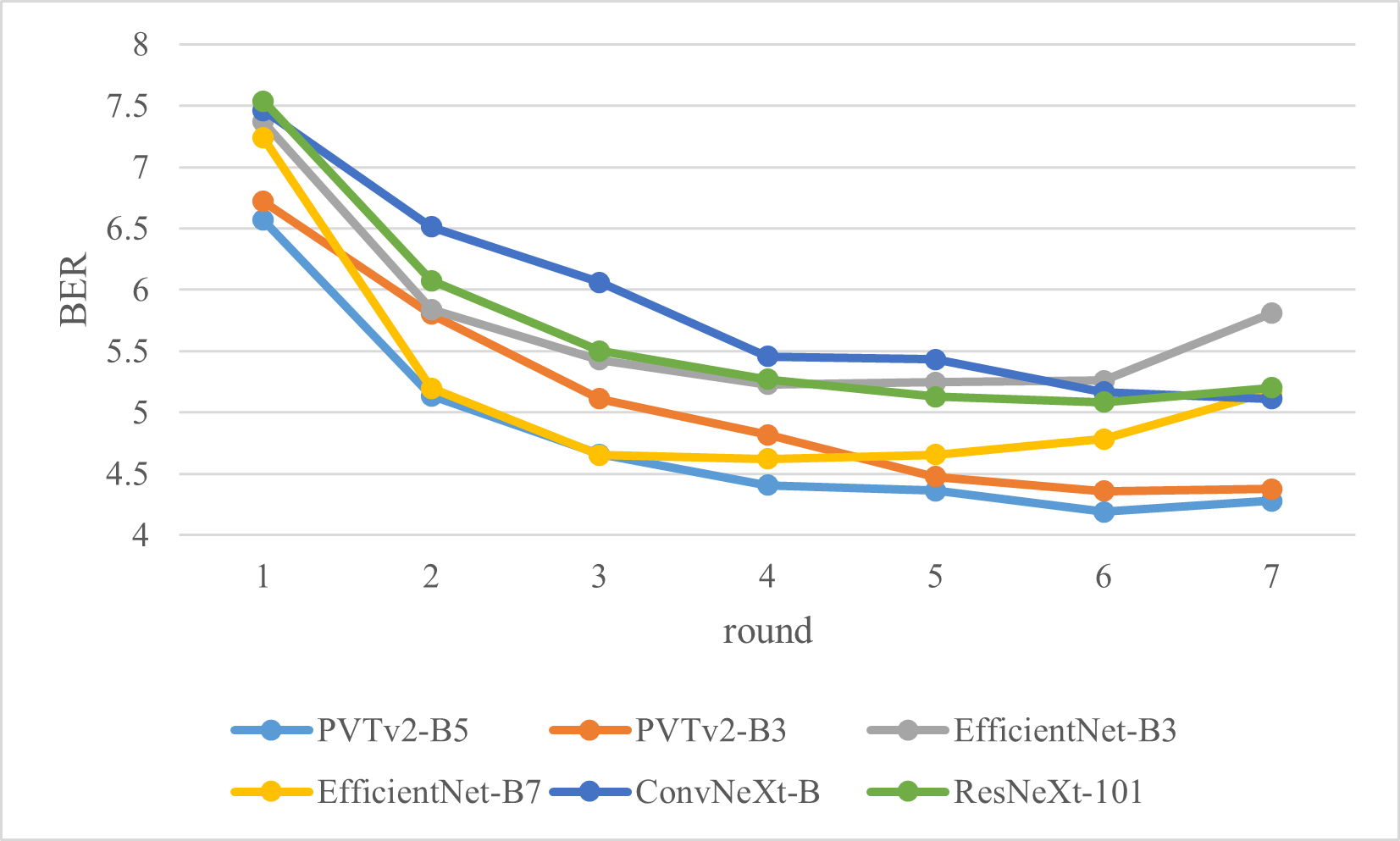}
    \caption{BER after each round of self-training on SBU.}
    \label{fig:BER_change_graph}
\end{figure}
}\fi


To show the effectiveness of our SILT, we further retrained the previous state-of-the-art networks on our tuned SBU and UCF dataset. The comparison results are shown in Table~\ref{comp_performance_improvement}, where all the previous networks achieve significant improvements, especially in terms of $\text{BER}_S$. 
It proves that the performance of existing networks is limited by the noisy labels and our tuned labels help to train a better deep model than using the original labels. 
It also shows that our SILT framework is applicable to various kinds of network architectures, not limited to the U-Net.



\subsection{Ablation Study}
\subsubsection{Component Analysis}
We conduct the ablation study by applying our SILT on  SBU~\cite{vicente2016large} dataset and testing it on our refined SBU test set. In the experiments, we use a U-Net~\cite{ronneberger2015u} with PVT v2-B5~\cite{wang2022pvt} as the backbone. In total, we consider the following baseline networks:
    1) Base: directly train a U-Net~\cite{ronneberger2015u} on the noisy dataset. 
    2) ST: directly apply a self-training framework, where we train a network, use the trained network to relabel the training data, and repeat these two stages alternately.  
    3) ST + Filter: add shadow-aware filtering in label tuning, but without global-local fusion. 
    4) ST + Fusion: add global-local fusion in label tuning, but without shadow-aware filtering. 
    5) ST + LT: full label tuning but without shadow counterfeiting. 
    6) ST + LT + SDA: add the common strong data augmentations,~\eg, RandomPerspective, GaussianBlur, etc.  
    7) ST + LT + SDA + SC: further use shadow counterfeiting in the network tuning stage.  
    8) Full: further add an additional unlabeled non-shadow dataset to the training data.

\begin{table}[tp]
    \centering
    \caption{Component analysis on the proposed SILT.}
    \renewcommand\tabcolsep{45.0pt}
    \resizebox{\linewidth}{!}{
    \begin{tabular}{l|c}
        \toprule
        & \multicolumn{1}{l}{BER$\downarrow$} \\ \hline
        \begin{tabular}[l]{@{}l@{}}Base\\ ST\\ ST+Filter\\ST+Fusion\\ ST+LT\\ ST+LT+SDA\\ ST+LT+SDA+SC\end{tabular} & \begin{tabular}[c]{@{}c@{}}6.62\\ 5.82\\ 5.44\\ 4.67\\ 4.52\\ 4.46\\ 4.37\end{tabular} \\ \hline
        Full& \textbf{4.19} \\ \bottomrule
    \end{tabular}}
    \label{ablation_study}
\end{table}

\begin{figure}[tp]  
	\centering
	\begin{minipage}[t]{0.88 \linewidth}
	    \includegraphics[width=\linewidth]{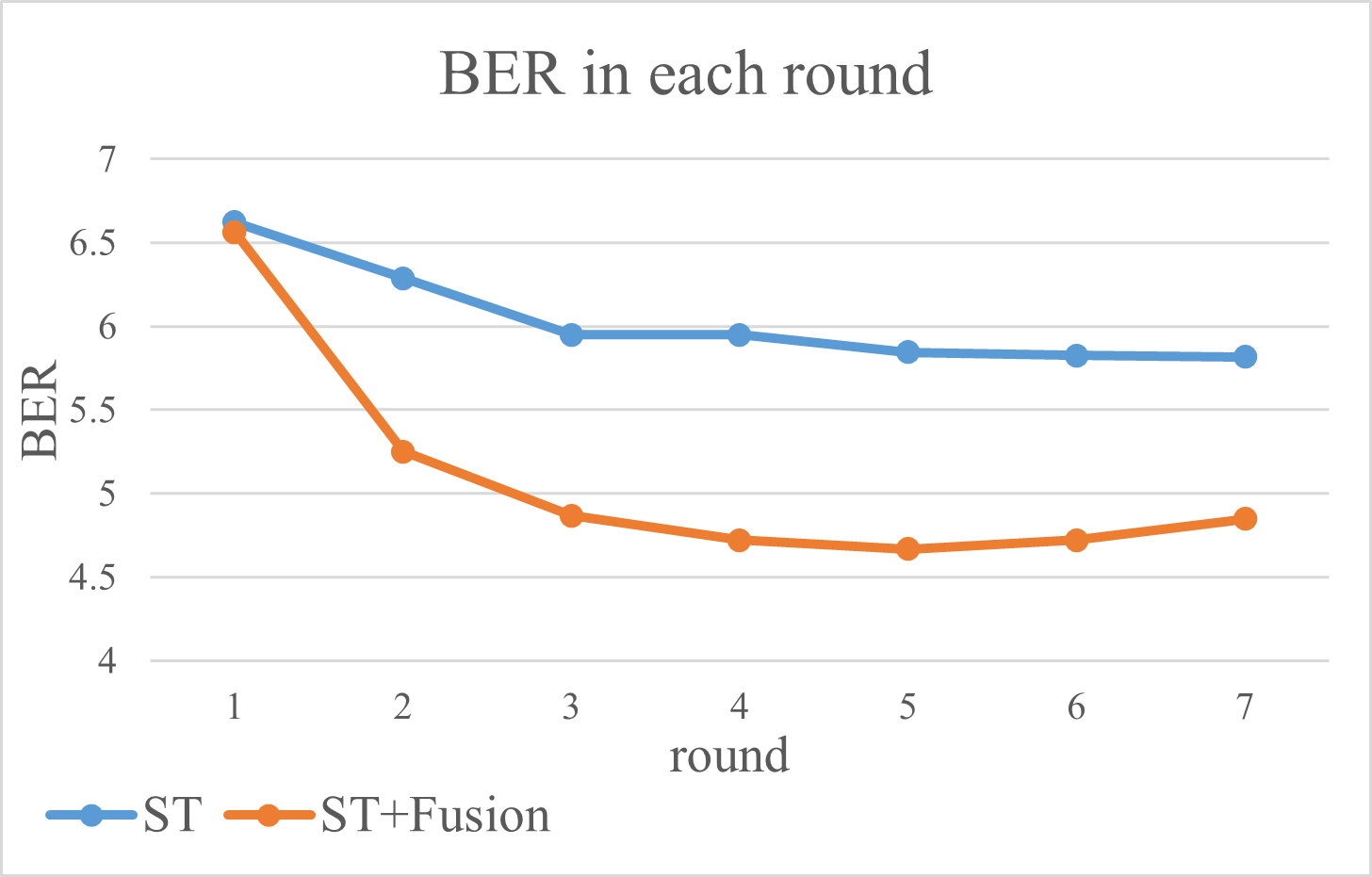}
	\end{minipage}
	\begin{minipage}[t]{0.88 \linewidth}
		\centerline{\footnotesize (a)}
	\end{minipage}
        \begin{minipage}[t]{0.88\linewidth}
		\includegraphics[width=\linewidth]{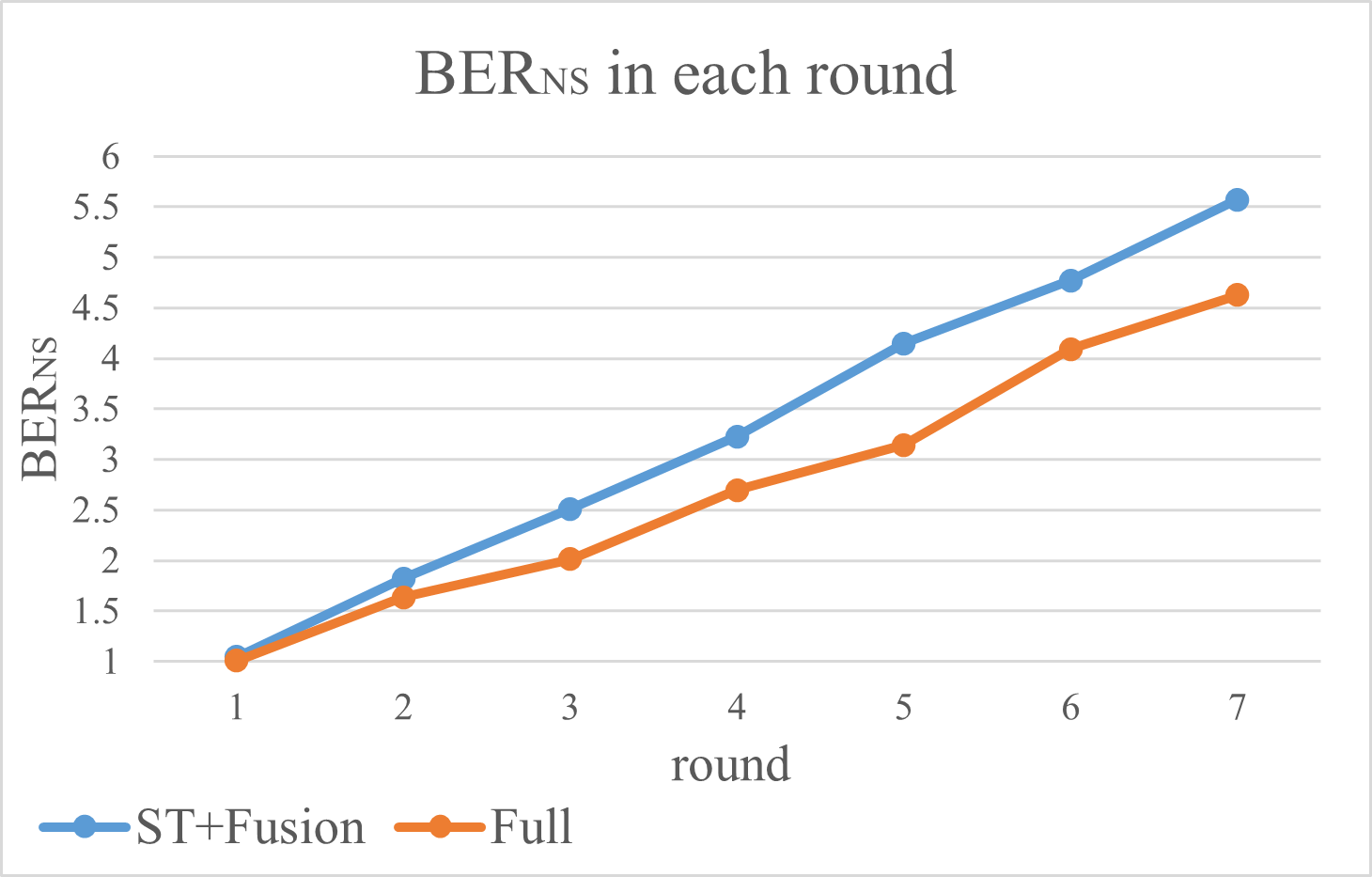}
	\end{minipage}
        \begin{minipage}[t]{0.88\linewidth}
		\centerline{\footnotesize (b)}
	\end{minipage}
	\caption{Comparison of the baseline networks in terms of error rates in each round.}
	\label{Fig: BER_comp}	
\end{figure}

Table ~\ref{ablation_study} shows the BER of each variant evaluated on our relabeled SBU test set, where we can see that 
(i) simply applying a self-training framework (ST) to refine noisy labels gives a limited improvement; (ii) each component in our framework design improves the quality of shadow masks in training data, thus promoting the performance of shadow detection; and (iii) our full pipeline with the additional non-shadow training set achieves the best performance, showing the effectiveness of the proposed method.

Table~\ref{ablation_study} shows that global-local fusion (ST + Fusion) brings the largest improvement. Fig.~\ref{Fig: BER_comp} (a) shows the BER in each round of the self-training framework with and without global-local fusion. 
Compared with ST, the ST + Fusion effectively alleviates overfitting in the first few training rounds. 
Fig.~\ref{Fig: BER_comp} (b) shows that our full pipeline with shadow-aware filtering and shadow counterfeiting reliefs error accumulation in shadow regions.

\begin{figure}[tp]
    \centering
    \includegraphics[width=0.95\linewidth]{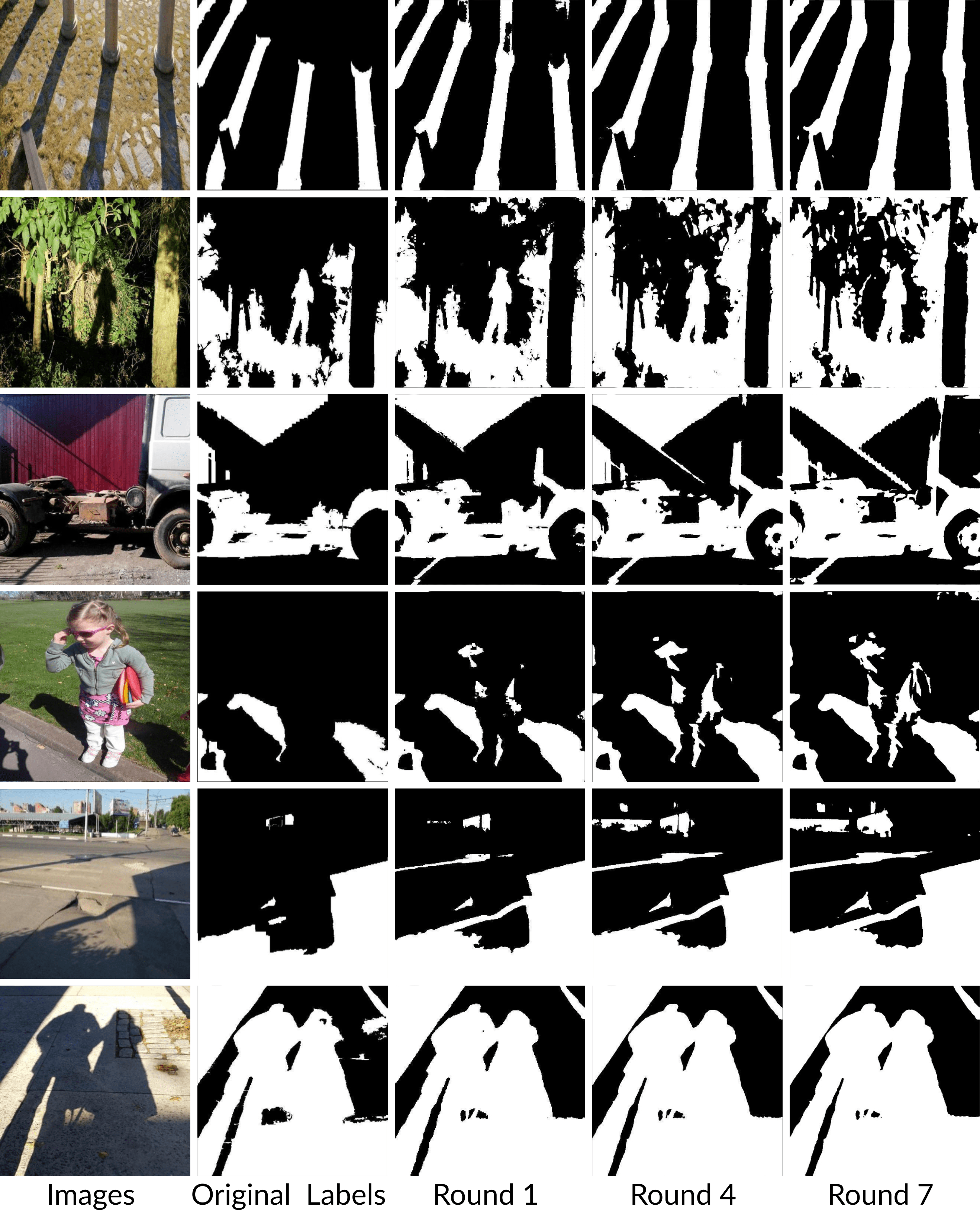}
    \caption{Shadow masks refined by different rounds of SILT.}
    \label{fig:shadow_masks}
\end{figure}

\subsubsection{Architecture Analysis}

\paragraph{Round number.} In Fig.~\ref{fig:shadow_masks}, we show the results where we use label tuning for different rounds in the training stage.
We can observe that (i) after the first round, we are able to relabel the large shadow regions in the images; (ii) fine details are gradually added in the shadow masks during the training process; and (iii) if using a large number of training rounds (the last column), the shadow masks become noisy again due to the error accumulation.
Hence, we empirically set the round number as seven.

\vspace{-2mm}
\paragraph{Hyper-parameters: $R_{min}$ and $R_{max}$.}
$R_{min}$ and $R_{max}$ are two hyper-parameters in equation~\eqref{eq:2}, which control the threshold to split the non-confident predictions into shadow and non-shadow regions. Table~\ref{rmin_and_rmax} shows the BER of different combinations, where when the difference between $R_{min}$ and $R_{max}$ is neither 0 nor too large, the shadow-aware filtering gives the best results, 
and we set $R_{min}$ and $R_{max}$ as 0.5 and 0.6 in experiments.

\begin{table}[tp]
\centering
\caption{BER values of different $R_{min}$ and $R_{max}$.}
\renewcommand\tabcolsep{10.0pt}
\resizebox{0.99\linewidth}{!}{
\begin{tabular}{cc|ccccc}
\hline
\multicolumn{2}{c|}{\multirow{2}{*}{}}           & \multicolumn{5}{c}{$R_{max}$}         \\ \cline{3-7} 
\multicolumn{2}{c|}{}                            & 0.4  & 0.5  & 0.6  & 0.7  & 0.8  \\ \hline
\multicolumn{1}{c|}{\multirow{3}{*}{$R_{min}$}} & 0.4 & 4.77 & 4.44 & 4.40    & 4.41    & -    \\
\multicolumn{1}{c|}{}                      & 0.5 & -    & 4.53 & \textbf{4.19} & 4.75 & -    \\
\multicolumn{1}{c|}{}                      & 0.6 & -    & -    & 4.45 & 4.36 & 4.80 \\ \hline
\end{tabular}
}
\label{rmin_and_rmax}
\end{table}

\if{
\begin{table}[tp]
\centering
\caption{BER values of different $R_{corr}$}
\resizebox{0.65\linewidth}{!}{
\begin{tabular}{c|ccccc}
\hline
$R_{corr}$ & 0.91 & 0.93 & 0.95 & 0.97 & 0.99 \\ \hline
BER   & 4.59 & 4.32 & \textbf{4.19} & 4.51 & 4.52 \\ \hline
\end{tabular}
}
\label{Rcorr}
\end{table}

\begin{table*}[ht]
\caption{Comparison results of the networks trained on manually relabeled SBU test set and SILT relabeled training set.}
\centering
\renewcommand\tabcolsep{12.0pt}
\resizebox{0.99\linewidth}{!}{
\begin{tabular}{c|ccc|ccc|ccc|ccc}
\hline
\multirow{2}{*}{\begin{tabular}[c]{@{}c@{}}Exp.\\ Time\end{tabular}} & \multicolumn{3}{c|}{SDCM(a)} & \multicolumn{3}{c|}{SDCM(b)} & \multicolumn{3}{c|}{SDCM(c)} & \multicolumn{3}{c}{ours} \\ \cline{2-13} 
 & BER    & $\text{BER}_S$   & $\text{BER}_{NS}$  & BER    & $\text{BER}_S$   & $\text{BER}_{NS}$  & BER    & $\text{BER}_S$  & $\text{BER}_{NS}$  & BER   & $\text{BER}_S$& $\text{BER}_{NS}$ \\ \hline
1                                                                    & 5.29   & 8.95     & 1.63     & 4.43   & 3.95     & 4.91     & 4.39   & 4.68     & 4.10      & 3.99  & 4.38   & 3.60     \\
2                                                                    & 5.44   & 7.33     & 3.55     & 5.27   & 4.48     & 6.06     & 5.73   & 5.21     & 6.25     & 4.34  & 3.54   & 5.15    \\
3                                                                    & 5.18   & 7.85     & 2.51     & 4.42   & 4.27     & 4.57     & 4.26   & 4.34     & 4.18     & 3.61  & 3.79   & 3.44    \\ \hline
\end{tabular}
}
\label{table:NonshadER}
\end{table*}
}\fi

\begin{table}[tp]
\centering
\caption{BER values of different $R_{corr}$}
\renewcommand\tabcolsep{12.0pt} 
\resizebox{0.99\linewidth}{!}{
\begin{tabular}{c|ccccc}
\hline
$R_{corr}$ & 0.91 & 0.93 & 0.95 & 0.97 & 0.99 \\ \hline
BER   & 4.59 & 4.32 & \textbf{4.19} & 4.51 & 4.52 \\ \hline
\end{tabular}
}
\label{Rcorr}
\end{table}

\vspace{-2mm}
\paragraph{Hyper-parameters: $R_{corr}$.} $R_{corr}$ controls the least ratio of the previous mask that is contained in the new mask. 
A larger $R_{corr}$ leads to fewer label corrections, while a smaller $R_{corr}$ introduces more wrong labels.
As shown in Table~\ref{Rcorr}, 
we empirically set $R_{corr}$ as 0.95.


\begin{table}[tp]
\centering
\caption{BER values of different $R_{filt}$}
\renewcommand\tabcolsep{12.0pt} 
\resizebox{0.99\linewidth}{!}{
\begin{tabular}{c|ccccc}
\hline
$R_{corr}$ & 0.05 & 0.075 & 0.1 & 0.125 & 0.15 \\ \hline
BER   & 4.70 & 4.47 & \textbf{4.19} & 4.42 & 4.62 \\ \hline
\end{tabular}
}
\label{Rfilt}
\vspace{-2mm}
\end{table}

\vspace{-2mm}
\paragraph{Hyper-parameters: $R_{filt}$.} 
$R_{filt}$ determines how much we take the local-view prediction into account. 
A smaller $R_{filt}$ keeps more local-view prediction, and vice versa.
As shown in Table~\ref{Rfilt}, 
we empirically set $R_{filt}$ as 0.1.


\subsubsection{Analysis on Additional Data} ~\label{subsec: add}

\begin{table}[h]
\centering
\begin{tabular}{c|ccc}
\hline
Dataset Name                                                      & SBU  & UCF  & ISTD \\ \hline
\begin{tabular}[c]{@{}c@{}}with additional dataset\end{tabular} & 4.19 & 7.60    & 1.18    \\ \hline
\begin{tabular}[c]{@{}c@{}}w/o additional dataset\end{tabular}  & 4.37 & 7.23 & 1.16 \\ \hline
\end{tabular}
\vspace{2mm}
\caption{The BER values of PVT v2-B5 based model trained on different datasets with and without additional dataset.}
\label{fig:ablation_additional_dataset}
\end{table}

For a fair comparison, we conduct the ablation study on the additional non-shadow data to test its effectiveness. The results of our PVT v2-B5 based model trained on the three datasets are listed in Table~\ref{fig:ablation_additional_dataset}. As the results of UCF and ISTD show, data with excessive hard negative cases would deteriorate the training, leading to a performance degradation. Meanwhile, the results on all the three datasets show that, without the additional dataset, SILT is still, or even more, competitive.

\subsection{Discussion on Non-Shadow Error Rate}
From Table~\ref{tabel:comp1}, we can observe that our method has a higher  Non-shadow Error Rate ($\text{BER}_{NS}$) than the prior ones. We hypothesize that a model trained on a well-labeled dataset tends to produce balanced $\text{BER}_S$ and $\text{BER}_{NS}$ values. In contrast, the previous training datasets with lots of missing masks lead the networks to predict fewer shadow regions, resulting in $\text{BER}_S$ much higher than $\text{BER}_{NS}$.

To validate our assumption, we randomly split the relabeled SBU test set into a new training set of 538 images and a new test set of 100 images. Then, we train the SOTA method SDCM~\cite{zhu2022single} on this new training set [denoted as SDCM(c)] and test it on the new test set. 
We also used this same test set to evaluate the performance of our SILT and SDCM~\cite{zhu2022single} that trained on the original SBU training set [denoted as SDCM(a)] and our SILT relabeled training set [denoted as SDCM(b)].
%
%
We conduct the experiments three times, and each time, we randomly select a new test set of the size of 100.

\begin{table}[tp]
\caption{Comparison results of the state-of-the-art (SOTA) method~\cite{zhu2022single} trained on three different training sets: (a) the original SBU training set, (b) our SILT relabeled training set, and (c) a subset of manually relabeled SBU test set (538 images). The evaluation was conducted on the rest of the relabeled SBU test set.}
\vspace{1mm}
\centering
\renewcommand\tabcolsep{7.0pt} 
\resizebox{0.99\linewidth}{!}{
\begin{tabular}{c|ccc|ccc}
\hline
\multirow{2}{*}{\begin{tabular}[c]{@{}c@{}}Exp.\\ Time\end{tabular}} & \multicolumn{3}{c|}{SDCM(a)} & \multicolumn{3}{c}{SDCM(b)} \\ \cline{2-7} 
& BER    & $\text{BER}_S$   & $\text{BER}_{NS}$  & BER    & $\text{BER}_S$  & $\text{BER}_{NS}$  \\ \hline
1 & 5.29   & 8.95     & 1.63     & 4.43   & 3.95    & 4.91     \\
2 & 5.44   & 7.33     & 3.55     & 5.27   & 4.48    & 6.06     \\
3 & 5.18   & 7.85     & 2.51     & 4.42   & 4.27    & 4.57     \\ \hline
AVG. & 5.30   & 8.04     & 2.56  & 4.70   & 4.23    & 5.18     \\ \hline
\multirow{2}{*}{} & \multicolumn{3}{c|}{SDCM(c)} & \multicolumn{3}{c}{Ours}    \\ \cline{2-7} 
 & BER    & $\text{BER}_S$   & $\text{BER}_{NS}$  & BER    & $\text{BER}_S$  & $\text{BER}_{NS}$  \\ \hline
1 & 4.39   & 4.68     & 4.10      & 3.99   & 4.38    & 3.60      \\
2 & 5.73   & 5.21     & 6.25     & 4.34   & 3.54    & 5.15     \\
3 & 4.26   & 4.34     & 4.18     & 3.61   & 3.79    & 3.44     \\ \hline
AVG. & 4.79   & 4.74     & 4.84  & 3.98   & 3.90    & 4.06     \\ \hline
\end{tabular}
}
\label{table:NonshadER}
\end{table}

The results are shown in Table~\ref{table:NonshadER}, where we can observe that 
(i) from (c), SDCM achieves lower BER and balanced $\text{BER}_S$ and $\text{BER}_{NS}$ with a smaller well-labeled training set, supporting our hypothesis;
%
(ii) the comparison among SDCM (a-c) shows that our SILT-relabeled training set improves shadow detection performance and indicates the high quality of our SILT-relabeled dataset;
\section{Conclusion}
\label{sec:conclusion}

We revisit the shadow detection task by considering noise in the shadow labels, and design SILT, a novel Shadow-aware Iterative Label Tuning framework, to enable effective model training on data with noisy labels.
Technically, we present a label tuning strategy to encourage the network to get rid of overfitting and try large refinement on the shadow labels.
Also, we design two shadow-specific data augmentation strategies, which add ``fake'' shadows into the training images to improve the ability of the network to distinguish the shadows and dark objects.
Considering the noise in existing datasets, we carefully relabel the test set of SBU~\cite{vicente2016large} for evaluation and conduct 
various experiments.
The experimental results show that with SLIT, even a simple network can achieve state-of-the-art performance on shadow detection.

\vspace{-2mm}
\paragraph{Limitations.}
Our SILT may fail to distinguish the shadow regions and dark non-shadow regions that have very similar colors and textures to the shadow regions.
This is a very challenging issue in the shadow detection task, which detects shadow regions from single images.
In the future, we aim to explore depth information to more accurately detect shadows in 3D space, since the shadows are usually projected on the 2D background. 

\section*{Acknowledgements}
This work was supported by Shenzhen Portion of Shenzhen-Hong Kong Science and Technology Innovation Cooperation Zone under HZQB-KCZYB-20200089, the project \#MMT-p2-21 of the Shun Hing Institute of Advanced Engineering, The Chinese University of Hong Kong, Hong Kong Research Grants
Council GRF Projects (CUHK 14201620 \& CUHK 14201321), the National Key R\&D Program of China (2022ZD0160100), and the Shanghai Committee of Science and Technology (21DZ1100100). 

{\small
\bibliographystyle{ieee_fullname}
\bibliography{egbib}

\begin{thebibliography}{10}\itemsep=-1pt

\bibitem{Affinity}
Affinity photo for {iPad}.
\newblock \url{https://affinity.serif.com/en-us/photo/ipad/}.

\bibitem{chen2015webly}
Xinlei Chen and Abhinav Gupta.
\newblock Webly supervised learning of convolutional networks.
\newblock In {\em IEEE International Conference on Computer Vision}, pages
  1431--1439, 2015.

\bibitem{chen2020multi}
Zhihao Chen, Lei Zhu, Liang Wan, Song Wang, Wei Feng, and Pheng-Ann Heng.
\newblock A multi-task mean teacher for semi-supervised shadow detection.
\newblock In {\em IEEE/CVF Conference on Computer Vision and Pattern
  Recognition}, pages 5611--5620, 2020.

\bibitem{cubuk2019randaugment}
Ekin~Dogus Cubuk, Barret Zoph, Jon Shlens, and Quoc Le.
\newblock {RandAugment}: Practical automated data augmentation with a reduced
  search space.
\newblock In {\em Advances in Neural Information Processing Systems},
  volume~33, pages 18613--18624, 2020.

\bibitem{deng2009imagenet}
Jia Deng, Wei Dong, Richard Socher, Li-Jia Li, Kai Li, and Fei-Fei Li.
\newblock Image{N}et: A large-scale hierarchical image database.
\newblock In {\em IEEE/CVF Conference on Computer Vision and Pattern
  Recognition}, pages 248--255, 2009.

\bibitem{ding2019argan}
Bin Ding, Chengjiang Long, Ling Zhang, and Chunxia Xiao.
\newblock {ARGAN}: Attentive recurrent generative adversarial network for
  shadow detection and removal.
\newblock In {\em IEEE International Conference on Computer Vision}, pages
  10213--10222, 2019.

\bibitem{fang2021robust}
Xianyong Fang, Xiaohao He, Linbo Wang, and Jianbing Shen.
\newblock Robust shadow detection by exploring effective shadow contexts.
\newblock In {\em Proceedings of the 29th ACM International Conference on
  Multimedia}, pages 2927--2935, 2021.

\bibitem{ghosh2017robust}
Aritra Ghosh, Himanshu Kumar, and P~Shanti Sastry.
\newblock Robust loss functions under label noise for deep neural networks.
\newblock In {\em AAAI Conference on Artificial Intelligence}, volume~31, 2017.

\bibitem{goldberger2016training}
Jacob Goldberger and Ehud Ben-Reuven.
\newblock Training deep neural-networks using a noise adaptation layer.
\newblock In {\em International Conference on Learning Representations}, 2016.

\bibitem{guo2011single}
Ruiqi Guo, Qieyun Dai, and Derek Hoiem.
\newblock Single-image shadow detection and removal using paired regions.
\newblock In {\em IEEE/CVF Conference on Computer Vision and Pattern
  Recognition}, pages 2033--2040, 2011.

\bibitem{han2018masking}
Bo Han, Jiangchao Yao, Gang Niu, Mingyuan Zhou, Ivor Tsang, Ya Zhang, and
  Masashi Sugiyama.
\newblock Masking: A new perspective of noisy supervision.
\newblock {\em Advances in Neural Information Processing Systems}, 31, 2018.

\bibitem{han2019deep}
Jiangfan Han, Ping Luo, and Xiaogang Wang.
\newblock Deep self-learning from noisy labels.
\newblock In {\em IEEE International Conference on Computer Vision}, pages
  5138--5147, 2019.

\bibitem{hou2019large}
Le Hou, Tomás F.~Yago Vicente, Minh Hoai, and Dimitris Samaras.
\newblock Large scale shadow annotation and detection using lazy annotation and
  stacked {CNNs}.
\newblock {\em IEEE Transactions on Pattern Analysis and Machine Intelligence},
  43(4):1337--1351, 2021.

\bibitem{hu2019direction}
Xiaowei Hu, Chi-Wing Fu, Lei Zhu, Jing Qin, and Pheng-Ann Heng.
\newblock Direction-aware spatial context features for shadow detection and
  removal.
\newblock {\em IEEE Transactions on Pattern Analysis and Machine Intelligence},
  42(11):2795--2808, 2019.

\bibitem{hu2021revisiting}
Xiaowei Hu, Tianyu Wang, Chi-Wing Fu, Yitong Jiang, Qiong Wang, and Pheng-Ann
  Heng.
\newblock Revisiting shadow detection: A new benchmark dataset for complex
  world.
\newblock {\em IEEE Transactions on Image Processing}, 30:1925--1934, 2021.

\bibitem{Hu_2018_CVPR}
Xiaowei Hu, Lei Zhu, Chi-Wing Fu, Jing Qin, and Pheng-Ann Heng.
\newblock Direction-aware spatial context features for shadow detection.
\newblock In {\em IEEE/CVF Conference on Computer Vision and Pattern
  Recognition}, pages 7454--7462, 2018.

\bibitem{huang2011characterizes}
Xiang Huang, Gang Hua, Jack Tumblin, and Lance Williams.
\newblock What characterizes a shadow boundary under the sun and sky?
\newblock In {\em IEEE International Conference on Computer Vision}, pages
  898--905, 2011.

\bibitem{jie2022rmlanet}
Leiping Jie and Hui Zhang.
\newblock {RMLAN}et: Random multi-level attention network for shadow detection.
\newblock In {\em IEEE International Conference on Multimedia and Expo}, pages
  1--6. IEEE, 2022.

\bibitem{jin2010towards}
Hang Jin and Yanming Feng.
\newblock Towards an automatic road lane marks extraction based on isodata
  segmentation and shadow detection from large-scale aerial images.
\newblock In {\em Proceedings of the 24th International Federation of
  Surveyors}, pages 1--12, 2010.

\bibitem{khan2014automatic}
Salman~Hameed Khan, Mohammed Bennamoun, Ferdous Sohel, and Roberto Togneri.
\newblock Automatic feature learning for robust shadow detection.
\newblock In {\em IEEE/CVF Conference on Computer Vision and Pattern
  Recognition}, pages 1939--1946, 2014.

\bibitem{adam}
Diederik~P. Kingma and Jimmy Ba.
\newblock Adam: {A} method for stochastic optimization.
\newblock In {\em International Conference on Learning Representations}, 2015.

\bibitem{lalonde2010detecting}
Jean-Fran{\c{c}}ois Lalonde, Alexei~A. Efros, and Srinivasa~G. Narasimhan.
\newblock Detecting ground shadows in outdoor consumer photographs.
\newblock In {\em European Conference on Computer Vision}, pages 322--335,
  2010.

\bibitem{le2018a+d}
Hieu Le, Tom{\'a}s F.~Yago Vicente, Vu Nguyen, Minh Hoai, and Dimitris Samaras.
\newblock {A+D} {N}et: Training a shadow detector with adversarial shadow
  attenuation.
\newblock In {\em European Conference on Computer Vision}, pages 662--678,
  2018.

\bibitem{lee2013pseudo}
Dong-Hyun Lee et~al.
\newblock Pseudo-label: The simple and efficient semi-supervised learning
  method for deep neural networks.
\newblock In {\em International Conference on Machine Learning}, volume~3, page
  896. Atlanta, 2013.

\bibitem{li2019learning}
Xinzhe Li, Qianru Sun, Yaoyao Liu, Qin Zhou, Shibao Zheng, Tat-Seng Chua, and
  Bernt Schiele.
\newblock Learning to self-train for semi-supervised few-shot classification.
\newblock {\em Advances in Neural Information Processing Systems}, 32, 2019.

\bibitem{li2019supervised}
Yinlin Li, Lihao Jia, Zidong Wang, Yang Qian, and Hong Qiao.
\newblock Un-supervised and semi-supervised hand segmentation in egocentric
  images with noisy label learning.
\newblock {\em Neurocomputing}, 334:11--24, 2019.

\bibitem{lin2014microsoft}
Tsung-Yi Lin, Michael Maire, Serge Belongie, James Hays, Pietro Perona, Deva
  Ramanan, Piotr Doll{\'a}r, and C.~Lawrence Zitnick.
\newblock Microsoft {COCO}: Common objects in context.
\newblock In {\em European Conference on Computer Vision}, pages 740--755,
  2014.

\bibitem{liu2022adaptive}
Sheng Liu, Kangning Liu, Weicheng Zhu, Yiqiu Shen, and Carlos Fernandez-Granda.
\newblock Adaptive early-learning correction for segmentation from noisy
  annotations.
\newblock In {\em IEEE/CVF Conference on Computer Vision and Pattern
  Recognition}, pages 2606--2616, 2022.

\bibitem{liu2022convnet}
Zhuang Liu, Hanzi Mao, Chao-Yuan Wu, Christoph Feichtenhofer, Trevor Darrell,
  and Saining Xie.
\newblock A convnet for the 2020s.
\newblock In {\em IEEE/CVF Conference on Computer Vision and Pattern
  Recognition}, pages 11976--11986, 2022.

\bibitem{luo2022deep}
Yaoru Luo, Guole Liu, Yuanhao Guo, and Ge Yang.
\newblock Deep neural networks learn meta-structures from noisy labels in
  semantic segmentation.
\newblock In {\em AAAI Conference on Artificial Intelligence}, volume~36, pages
  1908--1916, 2022.

\bibitem{lyu2019curriculum}
Yueming Lyu and Ivor~W Tsang.
\newblock Curriculum loss: Robust learning and generalization against label
  corruption.
\newblock {\em arXiv preprint arXiv:1905.10045}, 2019.

\bibitem{mendel2020semi}
Robert Mendel, Luis Antonio~de Souza, David Rauber, Joao~Paulo Papa, and
  Christoph Palm.
\newblock Semi-supervised segmentation based on error-correcting supervision.
\newblock In {\em European Conference on Computer Vision}, pages 141--157.
  Springer, 2020.

\bibitem{menon2019can}
Aditya~Krishna Menon, Ankit~Singh Rawat, Sashank~J Reddi, and Sanjiv Kumar.
\newblock Can gradient clipping mitigate label noise?
\newblock In {\em International Conference on Learning Representations}, 2019.

\bibitem{min2019two}
Shaobo Min, Xuejin Chen, Zheng-Jun Zha, Feng Wu, and Yongdong Zhang.
\newblock A two-stream mutual attention network for semi-supervised biomedical
  segmentation with noisy labels.
\newblock In {\em AAAI Conference on Artificial Intelligence}, volume~33, pages
  4578--4585, 2019.

\bibitem{najibi2019autofocus}
Mahyar Najibi, Bharat Singh, and Larry~S Davis.
\newblock Autofocus: Efficient multi-scale inference.
\newblock In {\em IEEE International Conference on Computer Vision}, pages
  9745--9755, 2019.

\bibitem{nguyen2017shadow}
Vu Nguyen, Tom{\'a}s F.~Yago Vicente, Maozheng Zhao, Minh Hoai, and Dimitris
  Samaras.
\newblock Shadow detection with conditional generative adversarial networks.
\newblock In {\em IEEE International Conference on Computer Vision}, pages
  4510--4518, 2017.

\bibitem{nigam2000analyzing}
Kamal Nigam and Rayid Ghani.
\newblock Analyzing the effectiveness and applicability of co-training.
\newblock In {\em Proceedings of the International Conference on Information
  and Knowledge Management}, pages 86--93, 2000.

\bibitem{ouali2020semi}
Yassine Ouali, C{\'e}line Hudelot, and Myriam Tami.
\newblock Semi-supervised semantic segmentation with cross-consistency
  training.
\newblock In {\em IEEE/CVF Conference on Computer Vision and Pattern
  Recognition}, pages 12674--12684, 2020.

\bibitem{panagopoulos2011illumination}
Alexandros Panagopoulos, Chaohui Wang, Dimitris Samaras, and Nikos Paragios.
\newblock Illumination estimation and cast shadow detection through a
  higher-order graphical model.
\newblock In {\em IEEE/CVF Conference on Computer Vision and Pattern
  Recognition}, pages 673--680, 2011.

\bibitem{patrini2017making}
Giorgio Patrini, Alessandro Rozza, Aditya Krishna~Menon, Richard Nock, and
  Lizhen Qu.
\newblock Making deep neural networks robust to label noise: A loss correction
  approach.
\newblock In {\em IEEE/CVF Conference on Computer Vision and Pattern
  Recognition}, pages 1944--1952, 2017.

\bibitem{ren2018learning}
Mengye Ren, Wenyuan Zeng, Bin Yang, and Raquel Urtasun.
\newblock Learning to reweight examples for robust deep learning.
\newblock In {\em International Conference on Machine Learning}, pages
  4334--4343, 2018.

\bibitem{ronneberger2015u}
Olaf Ronneberger, Philipp Fischer, and Thomas Brox.
\newblock U-net: Convolutional networks for biomedical image segmentation.
\newblock In {\em International Conference on Medical Image Computing and
  Computer-Assisted Intervention}, pages 234--241. Springer, 2015.

\bibitem{salvador2004cast}
Elena Salvador, Andrea Cavallaro, and Touradj Ebrahimi.
\newblock Cast shadow segmentation using invariant color features.
\newblock {\em Computer Vision and Image Understanding}, 95(2):238--259, 2004.

\bibitem{shen2015shadow}
Li Shen, Teck Wee~Chua, and Karianto Leman.
\newblock Shadow optimization from structured deep edge detection.
\newblock In {\em IEEE/CVF Conference on Computer Vision and Pattern
  Recognition}, pages 2067--2074, 2015.

\bibitem{shu2019lvc}
Yucheng Shu, Xiao Wu, and Weisheng Li.
\newblock {LVC-N}et: Medical image segmentation with noisy label based on local
  visual cues.
\newblock In {\em International Conference on Medical Image Computing and
  Computer-Assisted Intervention}, pages 558--566. Springer, 2019.

\bibitem{tan2019efficientnet}
Mingxing Tan and Quoc Le.
\newblock Efficientnet: Rethinking model scaling for convolutional neural
  networks.
\newblock In {\em International Conference on Machine Learning}, pages
  6105--6114, 2019.

\bibitem{tanaka2018joint}
Daiki Tanaka, Daiki Ikami, Toshihiko Yamasaki, and Kiyoharu Aizawa.
\newblock Joint optimization framework for learning with noisy labels.
\newblock In {\em IEEE/CVF Conference on Computer Vision and Pattern
  Recognition}, pages 5552--5560, 2018.

\bibitem{tanno2019learning}
Ryutaro Tanno, Ardavan Saeedi, Swami Sankaranarayanan, Daniel~C Alexander, and
  Nathan Silberman.
\newblock Learning from noisy labels by regularized estimation of annotator
  confusion.
\newblock In {\em IEEE/CVF Conference on Computer Vision and Pattern
  Recognition}, pages 11244--11253, 2019.

\bibitem{tian2016new}
Jiandong Tian, Xiaojun Qi, Liangqiong Qu, and Yandong Tang.
\newblock New spectrum ratio properties and features for shadow detection.
\newblock {\em Pattern Recognition}, 51:85--96, 2016.

\bibitem{vahdat2017toward}
Arash Vahdat.
\newblock Toward robustness against label noise in training deep discriminative
  neural networks.
\newblock {\em Advances in Neural Information Processing Systems}, 30, 2017.

\bibitem{veit2017learning}
Andreas Veit, Neil Alldrin, Gal Chechik, Ivan Krasin, Abhinav Gupta, and Serge
  Belongie.
\newblock Learning from noisy large-scale datasets with minimal supervision.
\newblock In {\em IEEE/CVF Conference on Computer Vision and Pattern
  Recognition}, pages 839--847, 2017.

\bibitem{vicente2015leave}
Tom{\'a}s F.~Yago Vicente, Minh Hoai, and Dimitris Samaras.
\newblock Leave-one-out kernel optimization for shadow detection.
\newblock In {\em IEEE International Conference on Computer Vision}, pages
  3388--3396, 2015.

\bibitem{vicente2016noisy}
Tom{\'a}s F.~Yago Vicente, Minh Hoai, and Dimitris Samaras.
\newblock Noisy label recovery for shadow detection in unfamiliar domains.
\newblock In {\em IEEE/CVF Conference on Computer Vision and Pattern
  Recognition}, pages 3783--3792, 2016.

\bibitem{vicente2018leave}
Tom{\'a}s F.~Yago Vicente, Minh Hoai, and Dimitris Samaras.
\newblock Leave-one-out kernel optimization for shadow detection and removal.
\newblock {\em IEEE Transactions on Pattern Analysis and Machine Intelligence},
  40(3):682--695, 2018.

\bibitem{vicente2016large}
Tom{\'a}s F.~Yago Vicente, Le Hou, Chen-Ping Yu, Minh Hoai, and Dimitris
  Samaras.
\newblock Large-scale training of shadow detectors with noisily-annotated
  shadow examples.
\newblock In {\em European Conference on Computer Vision}, pages 816--832,
  2016.

\bibitem{wang2018stacked}
Jifeng Wang, Xiang Li, and Jian Yang.
\newblock Stacked conditional generative adversarial networks for jointly
  learning shadow detection and shadow removal.
\newblock In {\em IEEE/CVF Conference on Computer Vision and Pattern
  Recognition}, pages 1788--1797, 2018.

\bibitem{Wang_2021_CVPR}
Tianyu Wang, Xiaowei Hu, Chi-Wing Fu, and Pheng-Ann Heng.
\newblock Single-stage instance shadow detection with bidirectional relation
  learning.
\newblock In {\em IEEE/CVF Conference on Computer Vision and Pattern
  Recognition}, pages 1--11, June 2021.

\bibitem{Wang_2022_TPAMI}
Tianyu Wang, Xiaowei Hu, Pheng-Ann Heng, and Chi-Wing Fu.
\newblock Instance shadow detection with a single-stage detector.
\newblock {\em IEEE Transactions on Pattern Analysis and Machine Intelligence},
  pages 1--14, 2022.

\bibitem{Wang_2020_CVPR}
Tianyu Wang, Xiaowei Hu, Qiong Wang, Pheng-Ann Heng, and Chi-Wing Fu.
\newblock Instance shadow detection.
\newblock In {\em IEEE/CVF Conference on Computer Vision and Pattern
  Recognition}, June 2020.

\bibitem{wang2022pvt}
Wenhai Wang, Enze Xie, Xiang Li, Deng-Ping Fan, Kaitao Song, Ding Liang, Tong
  Lu, Ping Luo, and Ling Shao.
\newblock {PVT} v2: Improved baselines with pyramid vision transformer.
\newblock {\em Computational Visual Media}, 8(3):415--424, 2022.

\bibitem{wang2019symmetric}
Yisen Wang, Xingjun Ma, Zaiyi Chen, Yuan Luo, Jinfeng Yi, and James Bailey.
\newblock Symmetric cross entropy for robust learning with noisy labels.
\newblock In {\em IEEE International Conference on Computer Vision}, pages
  322--330, 2019.

\bibitem{wu2022light}
Wen Wu, Kai Zhou, Xiao-Diao Chen, and Jun-Hai Yong.
\newblock Light-weight shadow detection via gcn-based annotation strategy and
  knowledge distillation.
\newblock {\em Computer Vision and Image Understanding}, 216:103341, 2022.

\bibitem{xia2020robust}
Xiaobo Xia, Tongliang Liu, Bo Han, Chen Gong, Nannan Wang, Zongyuan Ge, and Yi
  Chang.
\newblock Robust early-learning: Hindering the memorization of noisy labels.
\newblock In {\em International Conference on Learning Representations}, 2020.

\bibitem{xiao2015learning}
Tong Xiao, Tian Xia, Yi Yang, Chang Huang, and Xiaogang Wang.
\newblock Learning from massive noisy labeled data for image classification.
\newblock In {\em IEEE/CVF Conference on Computer Vision and Pattern
  Recognition}, pages 2691--2699, 2015.

\bibitem{xie2020self}
Qizhe Xie, Minh-Thang Luong, Eduard Hovy, and Quoc~V Le.
\newblock Self-training with noisy student improves imagenet classification.
\newblock In {\em IEEE/CVF Conference on Computer Vision and Pattern
  Recognition}, pages 10687--10698, 2020.

\bibitem{xie2017aggregated}
Saining Xie, Ross Girshick, Piotr Doll{\'a}r, Zhuowen Tu, and Kaiming He.
\newblock Aggregated residual transformations for deep neural networks.
\newblock In {\em IEEE/CVF Conference on Computer Vision and Pattern
  Recognition}, pages 1492--1500, 2017.

\bibitem{yarowsky1995unsupervised}
David Yarowsky.
\newblock Unsupervised word sense disambiguation rivaling supervised methods.
\newblock In {\em 33rd Annual Meeting of the Association for Computational
  Linguistics}, pages 189--196, 1995.

\bibitem{yi2019probabilistic}
Kun Yi and Jianxin Wu.
\newblock Probabilistic end-to-end noise correction for learning with noisy
  labels.
\newblock In {\em IEEE/CVF Conference on Computer Vision and Pattern
  Recognition}, pages 7017--7025, 2019.

\bibitem{yuan2021simple}
Jianlong Yuan, Yifan Liu, Chunhua Shen, Zhibin Wang, and Hao Li.
\newblock A simple baseline for semi-supervised semantic segmentation with
  strong data augmentation.
\newblock In {\em IEEE International Conference on Computer Vision}, pages
  8229--8238, 2021.

\bibitem{zhang2020disentangling}
Le Zhang, Ryutaro Tanno, Mou-Cheng Xu, Chen Jin, Joseph Jacob, Olga Cicarrelli,
  Frederik Barkhof, and Daniel Alexander.
\newblock Disentangling human error from ground truth in segmentation of
  medical images.
\newblock {\em Advances in Neural Information Processing Systems},
  33:15750--15762, 2020.

\bibitem{zhang2020characterizing}
Minqing Zhang, Jiantao Gao, Zhen Lyu, Weibing Zhao, Qin Wang, Weizhen Ding,
  Sheng Wang, Zhen Li, and Shuguang Cui.
\newblock Characterizing label errors: confident learning for noisy-labeled
  image segmentation.
\newblock In {\em International Conference on Medical Image Computing and
  Computer-Assisted Intervention}, pages 721--730. Springer, 2020.

\bibitem{zhang2018generalized}
Zhilu Zhang and Mert Sabuncu.
\newblock Generalized cross entropy loss for training deep neural networks with
  noisy labels.
\newblock {\em Advances in Neural Information Processing Systems}, 31, 2018.

\bibitem{zheng2019distraction}
Quanlong Zheng, Xiaotian Qiao, Ying Cao, and Rynson~W.H. Lau.
\newblock Distraction-aware shadow detection.
\newblock In {\em IEEE/CVF Conference on Computer Vision and Pattern
  Recognition}, pages 5167--5176, 2019.

\bibitem{zhou2022shadow}
Kai Zhou, Wen Wu, Yan-Li Shao, Jing-Long Fang, Xing-Qi Wang, and Dan Wei.
\newblock Shadow detection via multi-scale feature fusion and unsupervised
  domain adaptation.
\newblock {\em Journal of Visual Communication and Image Representation},
  88:103596, 2022.

\bibitem{zhou2020robust}
Tianyi Zhou, Shengjie Wang, and Jeff Bilmes.
\newblock Robust curriculum learning: from clean label detection to noisy label
  self-correction.
\newblock In {\em International Conference on Learning Representations}, 2020.

\bibitem{zhou2021asymmetric}
Xiong Zhou, Xianming Liu, Junjun Jiang, Xin Gao, and Xiangyang Ji.
\newblock Asymmetric loss functions for learning with noisy labels.
\newblock In {\em International Conference on Machine Learning}, pages
  12846--12856, 2021.

\bibitem{zhou2021learning}
Xiong Zhou, Xianming Liu, Chenyang Wang, Deming Zhai, Junjun Jiang, and
  Xiangyang Ji.
\newblock Learning with noisy labels via sparse regularization.
\newblock In {\em IEEE International Conference on Computer Vision}, pages
  72--81, 2021.

\bibitem{zhu2010learning}
Jiejie Zhu, Kegan~G.G. Samuel, Syed~Z. Masood, and Marshall~F. Tappen.
\newblock Learning to recognize shadows in monochromatic natural images.
\newblock In {\em IEEE/CVF Conference on Computer Vision and Pattern
  Recognition}, pages 223--230, 2010.

\bibitem{zhu2018bidirectional}
Lei Zhu, Zijun Deng, Xiaowei Hu, Chi-Wing Fu, Xuemiao Xu, Jing Qin, and
  Pheng-Ann Heng.
\newblock Bidirectional feature pyramid network with recurrent attention
  residual modules for shadow detection.
\newblock In {\em European Conference on Computer Vision}, pages 121--136,
  2018.

\bibitem{zhu2021mitigating}
Lei Zhu, Ke Xu, Zhanghan Ke, and Rynson~WH Lau.
\newblock Mitigating intensity bias in shadow detection via feature
  decomposition and reweighting.
\newblock In {\em IEEE International Conference on Computer Vision}, pages
  4702--4711, 2021.

\bibitem{zhu2022single}
Yurui Zhu, Xueyang Fu, Chengzhi Cao, Xi Wang, Qibin Sun, and Zheng-Jun Zha.
\newblock Single image shadow detection via complementary mechanism.
\newblock In {\em Proceedings of the 30th ACM International Conference on
  Multimedia}, pages 6717--6726, 2022.

\end{thebibliography}
}

\end{document}